\documentclass{article}

\usepackage{PRIMEarxiv}

\usepackage[utf8]{inputenc} 
\usepackage[T1]{fontenc}    
\usepackage{hyperref}       
\usepackage{url}            
\usepackage{booktabs}       
\usepackage{amsfonts}       
\usepackage{nicefrac}       
\usepackage{subfig}   
\usepackage{multirow} 
\usepackage{amsmath} 
\usepackage{fancyhdr}       
\usepackage{graphicx}       
\usepackage{algorithmic}
\usepackage{algorithm}
\usepackage{array}
\usepackage{textcomp}
\usepackage{stfloats}
\usepackage{verbatim}
\usepackage{cite}
\usepackage{comment}
\usepackage{float}
\usepackage[table]{xcolor}
\usepackage[normalem]{ulem}
\usepackage{colortbl}
\usepackage{amssymb}
\usepackage{tikz}

\graphicspath{{figures/}}     

\definecolor{background}{RGB}{0,0,0}
\definecolor{tree}{RGB}{255,0,0}
\definecolor{asphalt}{RGB}{0,255,0}
\definecolor{brick}{RGB}{0,0,255}
\definecolor{bitumen}{RGB}{255,255,0}
\definecolor{shadow}{RGB}{0,255,255}
\definecolor{meadow}{RGB}{255,0,255}
\definecolor{baresoil}{RGB}{128,0,0}
\definecolor{rocksandsand}{RGB}{0,128,0}
\definecolor{coastalwater}{RGB}{128,0,128}
\definecolor{unknown}{RGB}{255,255,255}

\definecolor{gamma01}{RGB}{31,119,180}   
\definecolor{gamma03}{RGB}{255,127,14}   
\definecolor{gamma05}{RGB}{44,160,44}    
\definecolor{gamma07}{RGB}{214,39,40}    
\definecolor{gamma09}{RGB}{148,103,189}  

\newcommand\crule[3][black]{\textcolor{#1}{\rule{#2}{#3}}}

\hypersetup{
  colorlinks=true,
  citecolor=blue,  
  linkcolor=blue,  
  urlcolor=blue,   
  filecolor=blue   
}

\DeclareGraphicsExtensions{.pdf,.png,.jpg,.svg}
\def\BibTeX{{\rm B\kern-.05em{\sc i\kern-.025em b}\kern-.08em
    T\kern-.1667em\lower.7ex\hbox{E}\kern-.125emX}}

\begin{document}

\title{Open-Set Domain Generalization through Spectral-Spatial Uncertainty Disentanglement for Hyperspectral Image Classification}

\author{
  Amirreza Khoshbakht \\
  Faculty of Engineering and Natural Sciences (VPALab),\\
  Sabanci University\\
  Orhanlı, Tuzla, 34956, İstanbul, Türkiye\\
  \texttt{amirreza.khoshbakht@sabanciuniv.edu} \\
   \And
  Erchan Aptoula \\
  Faculty of Engineering and Natural Sciences (VPALab), \\
  Sabanci University\\
  Orhanlı, Tuzla, 34956, İstanbul, Türkiye\\
  \texttt{erchan@sabanciuniv.edu} \\
}

\maketitle

\begin{abstract}
Open-set domain generalization (OSDG) tackles the dual challenge of recognizing unknown classes while simultaneously striving to generalize across unseen domains without using target data during training. 
In this article, an OSDG framework for hyperspectral image classification is proposed, centered on a new Spectral–Spatial Uncertainty Disentanglement mechanism. It has been designed to address the domain shift influencing both spectral, spatial and combined feature extraction pathways 
using evidential deep learning, after which the most reliable pathway for each sample is adaptively selected. The proposed framework is further integrated with frequency-domain feature extraction for domain-invariant representation learning, dual-channel residual networks for spectral–spatial feature extraction, and evidential deep learning based uncertainty quantification. Experiments conducted on three cross-scene hyperspectral datasets, show that performance comparable to state-of-the-art domain adaptation methods can be achieved despite no access to target data, while high unknown-class rejection and known-class accuracy levels are maintained. The implementation will be available at github.com/amir-khb/UGOSDG upon acceptance.
\end{abstract}


\section{Introduction}
\label{sec1}
Hyperspectral image (HSI) classification constitutes a fundamental task in remote sensing applications such as precision agriculture, environmental monitoring, geological exploration, and urban planning \cite{paoletti2019deep}. However, practical deployment faces significant challenges when applied across different geographical locations, temporal periods, or sensor configurations, where performance-wise disruptive distribution shifts (known as ``domain shift'') occur commonly due to varying atmospheric conditions and sensor characteristics \cite{luo2022cross}.

Domain adaptation (DA) approaches address cross-domain HSI classification by typically leveraging labeled source data alongside limited target samples during training \cite{zhang2023locality}, assuming shared label spaces and target data availability. These assumptions are often violated in real-world scenarios where target domains may contain unknown classes and obtaining labeled target data can be expensive or even infeasible \cite{ma2024transfer}, leading to negative transfer phenomena where models incorrectly map unknown samples to known classes \cite{saito2018open}. Open-set domain adaptation (OSDA) on the other hand, addresses unknown classes during cross-domain transfer, with recent methods like WGDT \cite{bi2025wgdt} employing weighted adversarial networks and dynamic thresholding. However, these approaches still require unlabeled target data during training. 

Domain generalization (DG) on the other hand, represents a more practical paradigm, aiming to learn domain-invariant representations via only source domains \cite{peng2024disentanglement}. Although recent DG methods show promise through disentangled representation learning \cite{peng2024disentanglement}, invariant semantic DG shuffle networks \cite{gao2025invariant}, and language-guided dual-branch Mamba \cite{jin2025ldbmamba}, they primarily address closed-set scenarios. The combination of open-set recognition with DG (OSDG), aims to tackle the joint challenges of each sub-problem (Table \ref{tab:settings_comparison}) where models must handle the compound uncertainty arising from both unknown classes and distribution shifts without access to target domain data during training. Moreover, in the particular case of HSI, unknown classes can exhibit domain-dependent spectral signatures, thus complicating the learning of generalizable rejection criteria. Overall, OSDG remains largely unexplored in HSI despite its importance for real-world deployment.

The OSDG framework proposed in this article has been designed specifically for HSI classification. Its main novelty in order to address unknown class detection, consists in dynamically and adaptively selecting the most reliable feature extraction pathway for a given data sample. This is achieved through a new Spectral-Spatial Uncertainty Disentanglement (SSUD) network component, which relies on evidential deep learning \cite{sensoy2018evidential} to select the most reliable (in terms of predictive performance) of three possible feature extraction pathways: spectral, spatial and the combination thereof, thus promoting more robust decisions against unknown classes.

As far as domain shift is concerned, the proposed approach incorporates established techniques such as frequency-domain feature extraction \cite{he2024decoupled} to create domain-invariant representations, combined with the evidential uncertainty estimation from SSUD. Overall, the introduced OSDG strategy works by disentangling and evaluating uncertainties across these pathways, enabling the model to ``choose'' the best one on-the-fly and generalize effectively to new, unseen scenarios.

The contributions of this article can be summarized as: 1) This work presents the first OSDG framework for HSI classification; 2) introduces the SSUD component for evidential uncertainty based, reliability oriented spectral-spatial feature extraction pathway selection and 3) it is shown across three cross-scene hyperspectral datasets, that it achieves performance comparable to or surpassing state-of-the-art domain adaptation methods, despite operating under the significantly more challenging setting of having no access to target domain data during training.

\begin{table}[t]
\begin{center}
\caption{Domain shift and open-set approaches.}
\label{tab:settings_comparison}
\footnotesize
\begin{tabular}{p{4.0cm}p{1.8cm}p{1.5cm}}
\hline
\textbf{Method} & \textbf{Target Domain Access} & \textbf{Unknown Classes} \\
\hline
\textbf{Domain Adaptation} & \checkmark & \texttimes \\
\textbf{Domain Generalization} & \texttimes & \texttimes \\
\textbf{Open-Set Recognition} & \checkmark\textsuperscript{*} & \checkmark \\
\textbf{Open-Set Domain Adaptation} & \checkmark & \checkmark \\
\textbf{Open-Set Domain Generalization} & \texttimes & \checkmark \\
\hline
\end{tabular}
\\[0.2em]
\footnotesize \textsuperscript{*} Single domain only
\end{center}
\end{table}

\section{Related Works}\label{sec:related}
\subsection{Unsupervised Domain Adaptation for HSI Classification}
Unsupervised domain adaptation (UDA) has emerged to address cross-domain HSI classification challenges by leveraging labeled source domain data alongside unlabeled target domain samples during training \cite{ma2024transfer}. For example in \cite{zhang2023locality}, local manifold structure preservation while performing class-wise distribution adaptation has been explored, which helps maintain discriminative capabilities across different domains. Another promising direction utilizes self-supervised contrastive learning to enhance both feature discrimination and spectral information integration \cite{he2023selfsupervised}. Furthermore, recent work has explored separating domain-invariant and domain-specific features through decoupled contrastive learning, thereby improving cross-scene adaptation performance \cite{chen2024decoupled}. More recent developments have further advanced the field through several innovative techniques. For instance, pseudo-class distribution constraints have been used to guide multi-view feature alignment, improving the consistency of representations across different perspectives \cite{li2025pseudo}. Additionally, bidirectional adaptation frameworks have been proposed that employ triple-branch transformer architectures to simultaneously extract both domain-invariant and domain-specific features \cite{zhang2025bidirectional}. Moreover, recent work has integrated masked self-distillation with class-separable adversarial training to enhance feature discriminability and prevent misclassifications between ambiguous categories \cite{zhang2025masked}.

However, the aforementioned methods assume shared label spaces between domains and require target domain access during training, limiting applicability when unknown classes are present or target data is unavailable.

\subsection{Domain Generalization for HSI Classification} 
DG aims to develop models that perform effectively on unseen target domains by leveraging data from one or more source domains during training. In computer vision, DG typically involves multiple source domains to capture diverse variations and enhance robustness. However, in HSI classification, state-of-the-art methods \cite{peng2024disentanglement, gao2025invariant, jin2025ldbmamba, gao2024c3dg, wang2024two} often employ cross-dataset approaches, which are essentially single-source DG strategies, training on one dataset and evaluating on another to address domain shifts. Our proposed method aligns with this single source domain paradigm.

Several key approaches have been developed for DG with HSI; for instance one strategy focuses on extracting latent domain-invariant representations to mitigate spectral heterogeneity issues through Transformer-based style transfer and feature disentanglement \cite{peng2024disentanglement}. Another direction incorporates convergence and constrained-risk theories to improve model stability across unseen domains \cite{gao2024c3dg}. Additionally, two-stage domain alignment networks have been proposed that leverage simulated images and supervised contrastive learning to enhance generalization performance \cite{wang2024two}.

The integration of semantic information and causal reasoning has emerged as a further important direction. Explicit high-level semantic networks have been developed that leverage text-based semantic information to achieve effective image-text alignment \cite{wang2024explicit}. In parallel, causal invariance approaches work to separate class-related features from domain-related features, improving generalization by focusing on causal relationships \cite{chen2024causal}.

More recent methods incorporate advanced architectures to further enhance generalization capabilities. One notable direction employs GAN-based frameworks that utilize feature style covariance for safe style and content randomization, incorporating a spatial shuffling discriminator and dual sampling direct adversarial contrastive learning to preserve domain-invariant semantics \cite{gao2025invariant}. Additionally, language-guided dual-branch approaches have emerged that employ spatial local-global and spectral limited-board scans with a spatial-spectral-star-fusion inhibited Mamba module, which is further enhanced by contrastive learning using both label-based and text-based prior knowledge \cite{jin2025ldbmamba}.

While these methods show effective cross-domain generalization, they primarily address closed-set scenarios where all target classes are assumed present in the source domain.

\subsection{Open-Set Recognition and Domain Adaptation for HSI Classification}
Open-set recognition identifies unknown classes absent during training, addressing real-world HSI scenarios where novel land cover types emerge \cite{cevikalp2023anomaly}. Traditional closed-set methods assume all classes are known during training, which fails when encountering novel classes or land cover changes. Closed-set methods learn inter-class boundaries rather than intrinsic distributions, causing known class features to occupy entire latent spaces with no room for novel classes \cite{chen2023learning}. HSI-specific advances have introduced several innovative architectures. One approach employs hybrid attention networks that enable effective few-shot learning, allowing models to generalize from limited training samples \cite{huang2025few}. Furthermore, multilayer perceptron architectures have been developed that fuse RGB and coded aperture snapshot spectral imaging measurements, thereby providing enhanced cross-modal information integration \cite{cai2024mlp}.

For OSDA methods, several key approaches have been developed to address the challenge of unknown classes in the target domain. One strategy employs adversarial DA with multiple auxiliary classifiers and weighting modules to distinguish the likelihood of known versus unknown classes \cite{shermin2021adversarial}. Building on this, deep mutual learning techniques have been proposed that alternate between sample separation and distribution matching to improve discrimination \cite{chang2024mind}.
For HSI-specific applications, weighted GANs have been introduced that utilize dynamic thresholding and class anchors for metric space learning through instance-level weighted-domain adversarial learning \cite{bi2025wgdt}. Additionally, uncertainty quantification approaches leverage unknown-aware domain adversarial learning to address the ambiguity of class boundaries \cite{jang2022unknown}. To handle imbalanced target distributions, open-set moving-threshold estimation methods have been developed that employ gradual alignment strategies \cite{ru2024imbalanced}.

Furthermore, recent work has explored adjustment and alignment techniques through front-door causal adjustment to improve unknown class detection \cite{li2023adjustment}. Finally, source-free scenarios have been addressed through open-set source-free DA, enabling privacy-constrained geographical applications where source data cannot be accessed \cite{wan2024unveiling}.

Despite these advances, combining open-set recognition with DG for HSI remains unexplored, representing a critical gap for deployment where target domains are typically unavailable during training.

\subsection{Evidential Learning for HSI Classification}
Evidential learning provides a framework for uncertainty quantification by modeling predictive distributions using higher-order probability distributions, enabling simultaneous prediction and uncertainty estimation for robust open-set and DA scenarios. The foundational framework places Dirichlet distributions over class probabilities, treating network outputs as evidence to provide both predictive probabilities and epistemic uncertainty estimates suitable for identifying out-of-distribution samples \cite{sensoy2018evidential}. Evidential learning extends to multi-label classification, improving identification of unknown label combinations \cite{chen2022evidential}.

For evidential DA and DG methods, several key approaches have emerged. One strategy utilizes uncertainty estimates through evidential learning to weight the importance of source and target samples, enabling adaptive domain-level and instance-level knowledge aggregation across multiple source domains \cite{pei2024evidential}. Another direction employs evidential deep learning with bi-level optimization and max rebiased confidence learning to strategically sequence domains based on reliability assessment, ensuring more stable adaptation \cite{peng2024advancing}.

As far as evidential learning applied to HSI is concerned, one approach utilizes uncertainty quantification to identify spectral regions that are affected by atmospheric or sensor variations under distribution shift, enabling more robust classification \cite{zhang2023evidential}. Additionally, methods have been proposed that combine evidential learning with attention mechanisms to achieve unified uncertainty estimation across both spectral and spatial information, specifically designed for hyperspectral open-set recognition \cite{liu2024spectral}. Furthermore, the integration of meta-learning with evidential learning has been explored to facilitate few-shot DA in hyperspectral scenarios where labeled data is limited \cite{park2023meta}.

The proposed approach tackles OSDG by employing evidential learning in a disentangled way across spectral-spatial pathways, enabling uncertainty-guided open-set classification.

\section{Proposed method}
\label{sec:method}

\begin{figure*}[t]
    \begin{center}
    \includegraphics[width=1.0\textwidth]{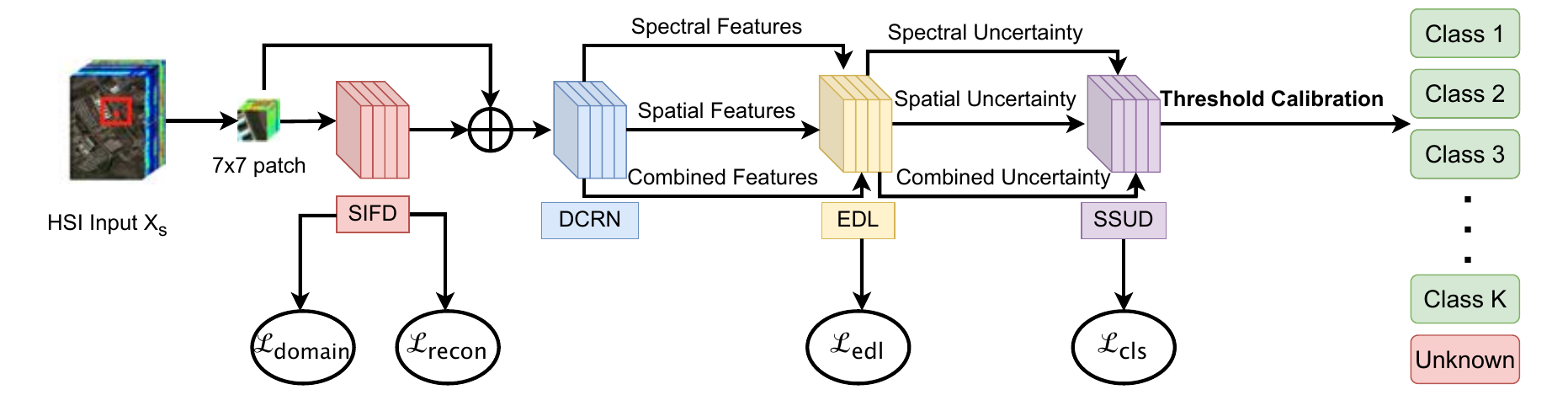}
    \caption{Illustration of the proposed OSDG framework with a HSI dataset of known $K$ classes. The architecture consists of four main components: SIFD for domain-invariant feature extraction, DCRN for spectral-spatial feature learning, EDL for uncertainty quantification, SSUD for uncertainty disentanglement, and finally threshold calibration before the open-set classification. The domain-invariant features from SIFD are combined with the original input via element-wise addition before being fed into DCRN. The loss functions $\mathcal{L}_{domain}$ (domain invariance loss), $\mathcal{L}_{recon}$ (reconstruction loss), $\mathcal{L}_{edl}$ (evidential deep learning loss), and $\mathcal{L}_{cls}$ (classification loss) are computed at their respective modules and combined for end-to-end training.}
    \label{fig:main_fig}
    \end{center}
\end{figure*}

\subsection{Architecture Overview}
The motivation behind the proposed framework rises from the fundamental challenge of OSDG, where models must simultaneously handle unknown classes and generalize across unseen domains without access to target domain data during training. This dual challenge requires a unified approach that can extract domain-invariant features while providing reliable uncertainty estimates to distinguish between known and unknown classes across different imaging conditions and geographical locations.

The proposed framework adopts a modular design where each of the four components specifically tackles a distinct aspect of the OSDG challenge:

\begin{enumerate}

    \item \textbf{Spectrum-Invariant Frequency Disentanglement (SIFD):} Addresses DG by extracting domain-agnostic spectral features in the frequency domain through attention-weighted feature selection and domain-agnostic regularization \cite{he2024decoupled}.

    \item \textbf{Dual-Channel Residual Network (DCRN):} Creates a suitable feature extractor for HSI by capturing complementary spectral-spatial information through parallel processing pathways combined using adaptive attention mechanisms \cite{bi2025wgdt}.
    
    \item \textbf{Evidential Deep Learning (EDL):} Enables effective open-set recognition through principled uncertainty quantification, where unknown classes naturally exhibit higher uncertainty than known classes \cite{sensoy2018evidential}.
    
    \item \textbf{Spectral-Spatial Uncertainty Disentanglement (SSUD):} is one of the novelties of this article, that leverages EDL's uncertainty estimates specifically for HSI by exploiting the different reliability of spatial and spectral pathways for more robust unknown detection.
\end{enumerate}

The overall data flow  (Fig.~\ref{fig:main_fig}) follows a sequential processing pipeline: the input HSI first undergoes domain-invariant feature extraction through SIFD, which operates in the frequency domain to suppress domain-specific variations while preserving essential spectral characteristics. These invariant features are then combined with the original input to create enhanced representations that maintain spectral fidelity while improving cross-domain robustness. The enhanced input is subsequently processed through DCRN's parallel spectral and spatial pathways to extract complementary feature representations. EDL then quantifies pathway-specific uncertainties using evidential learning principles, providing principled confidence estimates for each feature modality. Finally, SSUD performs open-set classification by adaptively weighting and disentangling spectral-spatial uncertainties to make reliable decisions about known versus unknown classes.


\begin{figure}[h]
  \begin{center}
  \includegraphics[width=0.5\textwidth]{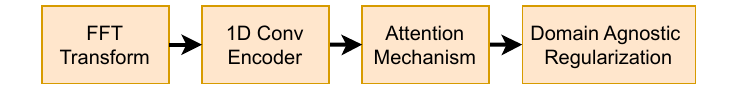}
  \caption{Outline of the SIFD component \cite{he2024decoupled}.}
  \label{fig.sifd}
\end{center}
\end{figure}

\subsection{Spectrum-Invariant Frequency Disentanglement (SIFD)}
HSIs from different domains exhibit domain-specific spectral variations due to atmospheric conditions, sensor differences, and environmental factors. The key insight behind SIFD is that while spectral signatures in the spatial domain are heavily influenced by domain-specific factors, their frequency domain representations reveal domain-invariant material characteristics that enable generalization to unseen domains without requiring synthetic domain generation~\cite{he2024decoupled}. Specifically, low-frequency components typically encode fundamental material properties that remain stable across domains, while high-frequency components often contain domain-specific noise and acquisition artifacts \cite{dong2016frequency} (Fig.~\ref{fig.sifd}).

For each pixel signature $\mathbf{s} \in \mathbb{R}^C$ (where C represents the number of spectral bands) extracted from the input HSI, Fast Fourier Transform is applied and the real and imaginary components are concatenated to obtain frequency features, which are then processed through a 1D convolutional encoder to extract frequency representations  $\mathbf{f}_{freq}$.

To enforce domain invariance, an attention mechanism is employed that learns channel-wise weights to emphasize frequency components corresponding to generalizable spectral characteristics while suppressing domain-specific variations.

\begin{figure*}[t]
  \begin{center}
  \includegraphics[width=0.9\textwidth]{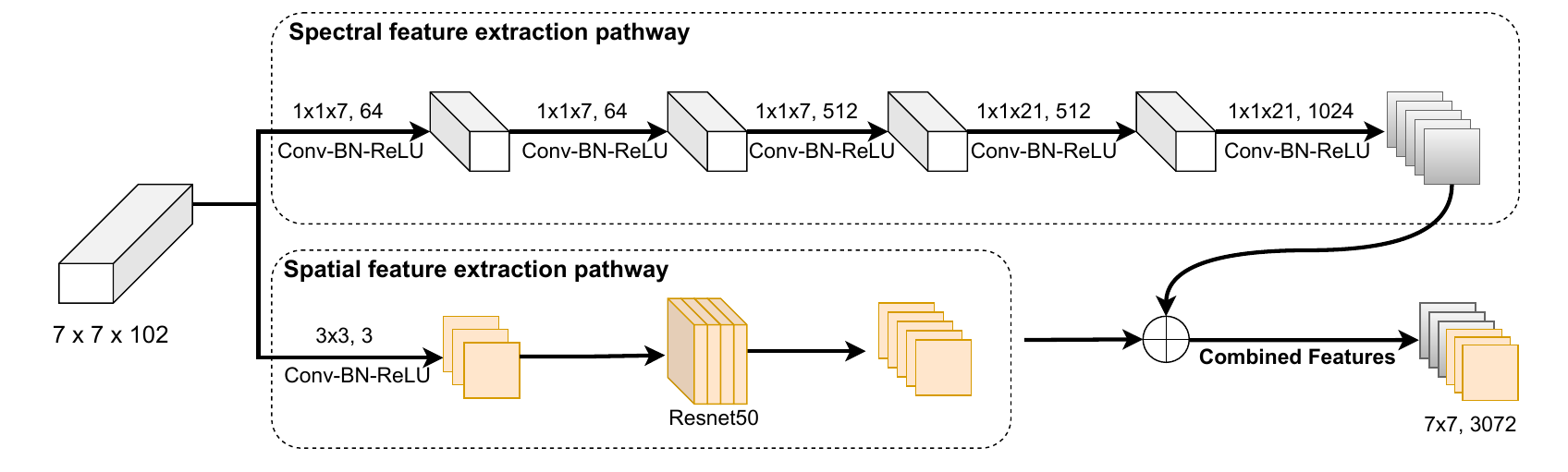} 
  \caption{DCRN processes hyperspectral patches through parallel spectral (2D convolutions) and spatial (ResNet50-based) pathways, fusing features via element-wise addition. The numbers in the blocks indicate the spatial and channel dimensions (height$\times$width$\times$channels) of feature maps throughout the convolutional layers.}
  \label{fig.dcrn}
\end{center}
\end{figure*}

Domain invariance is explicitly enforced through domain-agnostic regularization:
\begin{equation}
\mathcal{L}_{domain} = \text{BCE}(\text{MLP}(\text{GRL}(\mathbf{f}_{freq})), 0.5)
\end{equation}
where BCE denotes Binary Cross-Entropy loss, MLP denotes Multi-Layer Perceptron, and $\text{GRL}$ is a Gradient Reversal Layer. By optimizing toward a uniform domain probability of 0.5, the network learns to suppress domain-discriminative information, resulting in domain-invariant features. A reconstruction constraint $\mathcal{L}_{recon} = ||\mathbf{s} - \sigma(\text{MLP}(\mathbf{f}_{freq}))||_2^2$ ensures that domain-invariant features retain sufficient information for accurate classification, where $\sigma(\cdot)$ denotes the sigmoid activation function.

Finally, the domain-invariant features are combined with the original input via element-wise addition before being processed by the DCRN for spectral-spatial feature extraction.

\subsection{Dual-Channel Residual Network (DCRN)}
HSIs contain rich information in both spectral and spatial dimensions that provide complementary discriminative features. DCRN addresses this by designing two separate pathways for spectral and spatial feature extraction, followed by attention-based fusion; it has been widely used for hyperspectral feature extraction \cite{bi2025wgdt, sarpong2024dual, wang2025selective}.

DCRN extracts spectral features using 2D convolutions and spatial features through a ResNet50 backbone. These features are concatenated using sequential channel and spatial attention mechanisms with residual connections to produce the final combined features (Fig.~\ref{fig.dcrn}).

\subsection{Evidential Deep Learning (EDL) for Uncertainty Quantification}
Traditional open-set methods assume that unknown classes exhibit predictable uncertainty patterns relative to known classes. However, in cross-domain scenarios, unknown classes from shifted domains may exhibit uncertainty signatures that differ from source domain patterns due to distributional changes. EDL addresses this by modeling uncertainty through domain-invariant features extracted by DCRN, ensuring that uncertainty estimates remain reliable across domain boundaries.

For each pathway (i.e.~spectral, spatial, combined), first evidence $\mathbf{e}_i \in \mathbb{R}^K$ is generated:
\begin{equation}
\mathbf{e}_{i} = \max(0, \text{Linear}(\mathbf{f}_{i})) + \epsilon, \quad \boldsymbol{\alpha}_{i} = \mathbf{e}_{i} + 1
\end{equation}

\noindent where $\mathbf{f}_i$ represents the feature vector from DCRN for pathway $i$ (i.e., $\mathbf{f}_{\text{spec}}$, $\mathbf{f}_{\text{spat}}$, or $\mathbf{f}_{\text{comb}}$), $\epsilon$ is a small constant for numerical stability, $\boldsymbol{\alpha}_i \in \mathbb{R}^K$ are the Dirichlet distribution parameters, and $K$ is the total number of known classes.

Then, the uncertainty $u_i \in [0,1]$ for each pathway is computed as:
\begin{equation}
S_{i} = \sum_{k=1}^K \alpha_{i,k}, \quad
\mathbf{p}_{i} = \frac{\boldsymbol{\alpha}_{i}}{S_{i}}, \quad
u_{i} = \frac{K}{S_{i}}
\end{equation}
where higher values indicate more uncertainty and $S_i \in \mathbb{R}^+$ represents the Dirichlet strength (sum of parameters) and $\mathbf{p}_i \in [0,1]^K$ denotes the expected probability distribution over classes.

Next, the EDL loss combines mean squared error with regularization as follows:
\begin{equation}
\begin{aligned}
& \mathcal{L}_{MSE}  = \sum_{k=1}^K (\delta_{k,y} - p_{i,k})^2 \quad
\mathcal{L}_{reg}  = \sum_{k=1}^K (1 - \delta_{k,y}) \cdot \alpha_{i,k} \\
& \mathcal{L}_{EDL}  = \mathcal{L}_{MSE} + \lambda_{reg} \cdot \mathcal{L}_{reg}
\end{aligned}
\end{equation}
\noindent where $y \in \{1, \ldots, K\}$ is the true class label, $\delta_{k,y}$ represents the one-hot encoding of the true class, $\mathcal{L}_{MSE}$ is the mean squared error between predicted and true distributions, $\mathcal{L}_{reg}$ is the regularization term penalizing evidence for incorrect classes, and $\lambda_{reg}$ denotes the regularization weight.

\subsection{Spectral-Spatial Uncertainty Disentanglement (SSUD)}
Different feature modalities (spectral vs. spatial) may exhibit varying reliability across different samples and domains. In cross-domain scenarios, the reliability of spectral versus spatial pathways varies with domain characteristics. By performing reliability-based decoupling when pathway reliabilities differ significantly beyond a threshold, SSUD selectively relies on the more reliable pathway while downweighting the less reliable one, maintaining robust unknown detection even when domain shift differentially affects feature modalities. Note that EDL generates uncertainty estimates for each pathway independently (spectral, spatial, and combined) before any reliability assessment, enabling pathway-specific reliability scores to be computed as the inverse of uncertainty.

First, reliability scores are computed as the inverse of uncertainty:
\begin{equation}
r_{spec} = 1 - u_{spec}, \quad
r_{spat} = 1 - u_{spat}, \quad
\Delta r = |r_{spec} - r_{spat}|
\end{equation}
where $r_{spec}, r_{spat} \in [0,1]$ represent reliability scores for spectral and spatial pathways, and $\Delta r \in [0,1]$ denotes the reliability difference between pathways. When $\Delta r > \tau_{decouple}$, the system uses pathway-specific uncertainty:
\begin{equation}  
u_{final} = \begin{cases}
\max(u_{spec}, \kappa_{down} \cdot u_{spat}) & \text{if } r_{spec} > r_{spat} \\
\max(u_{spat}, \kappa_{down} \cdot u_{spec}) & \text{if } r_{spat} > r_{spec} \\
u_{combined} & \text{otherwise}
\end{cases}
\end{equation}
where $u_{final} \in [0,1]$ represents the final uncertainty measure, $\tau_{decouple}$ is the threshold for pathway decoupling, $\kappa_{down}$ denotes the down-weighting factor for less reliable pathway, and $u_{combined}$ is the uncertainty from the combined pathway.

Then the final prediction $\hat{y}$ including the unknown class is realized through a rejection score $r_{score}$ combining uncertainty and confidence measures as follows:
\begin{equation}
\begin{aligned}
&\mathbf{p}_{cls} = \text{Softmax}(\text{MLP}(\mathbf{f}_{comb})) \\
&p_{max} = \max(\mathbf{p}_{cls}), \quad \hat{k} = \arg\max(\mathbf{p}_{cls}) \\
&r_{score} = \omega_{unc} \cdot u_{final} + \omega_{conf} \cdot (1 - p_{max}) \\
&\hat{y} = \begin{cases}
\hat{k} & \text{if } r_{score} \leq \tau \\
\text{unknown} & \text{otherwise}
\end{cases}
\end{aligned}
\end{equation}
where $\mathbf{p}_{cls} \in [0,1]^K$ represents the class probability distribution and $\omega_{unc}$, $\omega_{conf}$ are empirically determined hyperparameters that balance uncertainty-based and confidence-based rejection criteria, selected through cross-validation on the source domain validation set and $\tau$ is the rejection threshold.

Finally, the complete training loss combines objectives from all components:
\begin{equation}
    \mathcal{L}_{total} = \mathcal{L}_{cls} + \alpha \mathcal{L}_{EDL} + \beta \mathcal{L}_{domain} + \gamma \mathcal{L}_{recon}
\end{equation}
where $\mathcal{L}_{cls} = \text{CrossEntropy}(\mathbf{p}_{cls}, y)$ is the classification loss, $\mathcal{L}_{EDL}$ is the evidential loss, $\mathcal{L}_{domain}$ is the domain-agnostic regularization, $\mathcal{L}_{recon}$ is the reconstruction constraint ensuring domain-invariant features retain spectral fidelity, and $\alpha$, $\beta$, $\gamma$ are weighting hyperparameters.

\subsection{Threshold Calibration via Synthetic Unknown Generation}
\label{sec.threshold}
Open-set classification performance critically depends on the rejection threshold, which determines the trade-off between correctly classifying known samples and rejecting unknown ones. Since true unknown samples from the target domain are unavailable during training, synthetic unknown generation provides a principled approach to calibrate thresholds \cite{ZHENG2024, du2022towards, tao2023nonparametric}.

Unknown sample generation has been achieved through \textit{Gaussian Noise Injection}:
\begin{equation}
\mathbf{x}_{noise} = \mathbf{x} + \mathcal{N}(0, \sigma^2 \mathbf{I})
\end{equation}
where $\sigma$ represents different noise levels; \textit{Class Mixing}:
\begin{equation}
\mathbf{x}_{mix} = \lambda \mathbf{x}_{i} + (1-\lambda) \mathbf{x}_{j}
\end{equation}
where $\lambda$ is the mixing weight and $\mathbf{x}_i, \mathbf{x}_j$ are samples from different classes; and \textit{Spectral and Spatial Corruption}, where random selected 20-30\% of spectral bands and 2-3 spatial patches are corrupted.

The optimal threshold $\tau^*$ is selected using ROC analysis to achieve a target unknown rejection rate $\rho_{target}$:
\begin{equation}
\tau^* = \arg\min_\tau |TPR(\tau) - \rho_{target}|
\end{equation}

This approach aims for robust performance across diverse hyperspectral domains while maintaining computational efficiency through focused feature extraction and uncertainty-guided decision making.

\section{Experiments}\label{sec:exp}
\subsection{Datasets}
In order to verify the effectiveness of the proposed method, six datasets are used: Pavia University (PU) \cite{pavia_datasets}, Pavia Center (PC) \cite{pavia_datasets}, Houston2013 (HU13) \cite{debes2014hyperspectral}, Houston2018 (HU18) \cite{roy2018multimodal}, Dioni, and Loukia \cite{hyrank2018}. They are combined into three OSDA classification tasks in pairs, where one dataset is used as the source domain and the other as a target domain: PU-PC, HU13-HU18 and Dioni-Loukia.
The selection of unknown classes for each dataset follows the standard protocols established in state-of-the-art DA literature \cite{bi2025wgdt}. 
For each source domain, a stratified split is employed where 80\% of the labeled samples are used for training and 20\% are reserved for validation.

\subsubsection{PU–PC}
\begin{figure}[!t]
\begin{center}
\includegraphics[width=1.0\columnwidth]{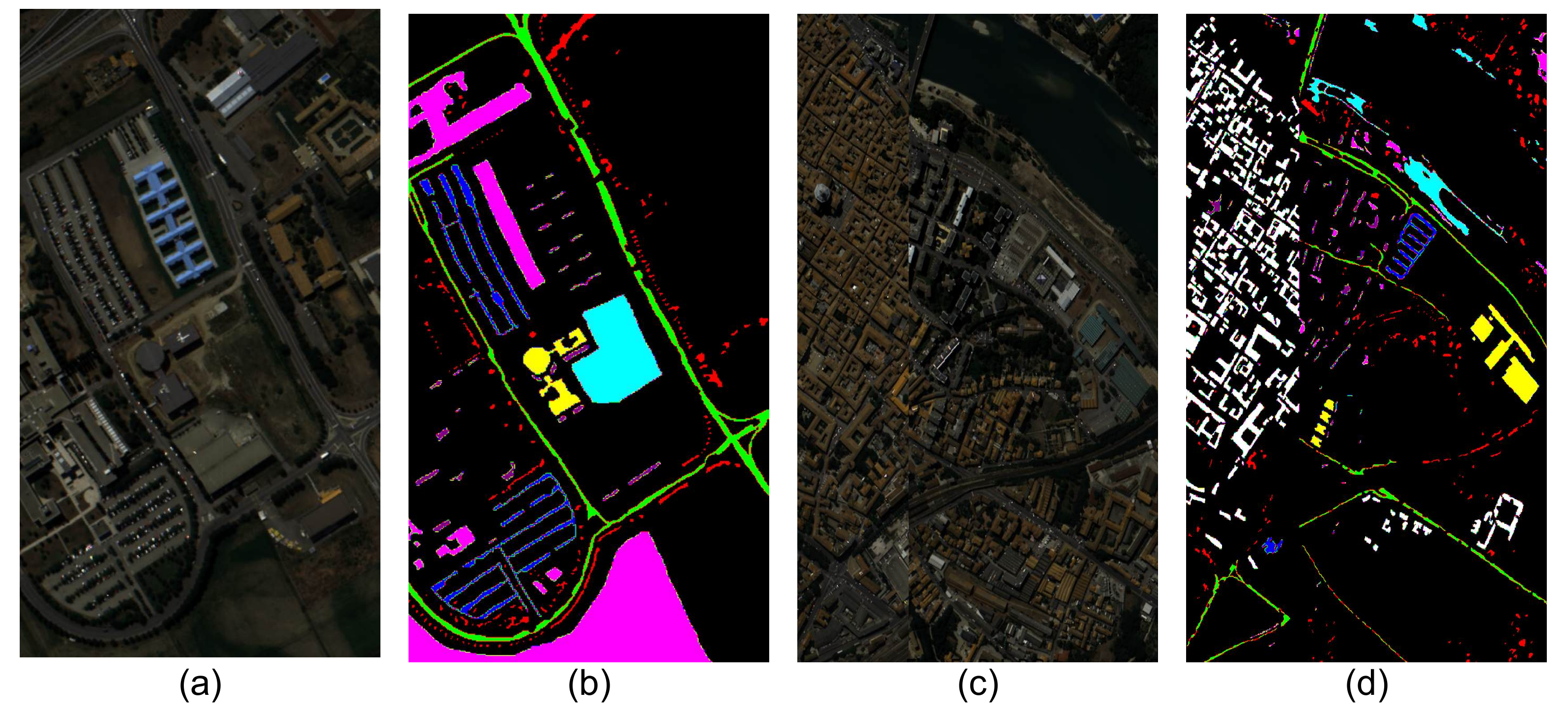}
\caption{PU–PC task. (a) Pseudo-color image of PU with 39,332 labeled pixels. (b) Ground-truth image of PU. (c) Pseudo-color image of PC with 82,181 labeled pixels. (d) Ground-truth image of PC. Class distribution (PU-PC): \crule[background]{0.5cm}{0.25cm} Background,
\crule[brick]{0.5cm}{0.25cm} Brick (3682-2685),
\crule[meadow]{0.5cm}{0.25cm} Meadow (18649-3090),
\crule[tree]{0.5cm}{0.25cm} Tree (3064-7598),
\crule[bitumen]{0.5cm}{0.25cm} Bitumen (1330-7287),
\crule[baresoil]{0.5cm}{0.25cm} Bare Soil (5029-6584),
\crule[asphalt]{0.5cm}{0.25cm} Asphalt (6631-9248),
\crule[shadow]{0.5cm}{0.25cm} Shadow (947-2863),
{\setlength{\fboxsep}{0pt}\fcolorbox{black}{unknown}{\rule{0pt}{0.25cm}\hspace{0.5cm}}} Unknown (0-42826)}
\label{fig:pu_pc}
\end{center}
\end{figure}

PU (610 $\times$ 340 pixels) and PC (1096 $\times$ 715 pixels), were acquired with the ROSIS sensor over the city of Pavia, Italy, and both have 1.3 m spatial resolution. After noise removal, 103 bands remain for PU and 102 for PC; experiments use 102 common bands. Seven common classes serve as known classes, with one class (Tiles) in the target domain set as unknown (Fig.~\ref{fig:pu_pc}).

\subsubsection{HU13–HU18}
\begin{figure}[!t]
\begin{center}
\includegraphics[width=0.7\columnwidth]{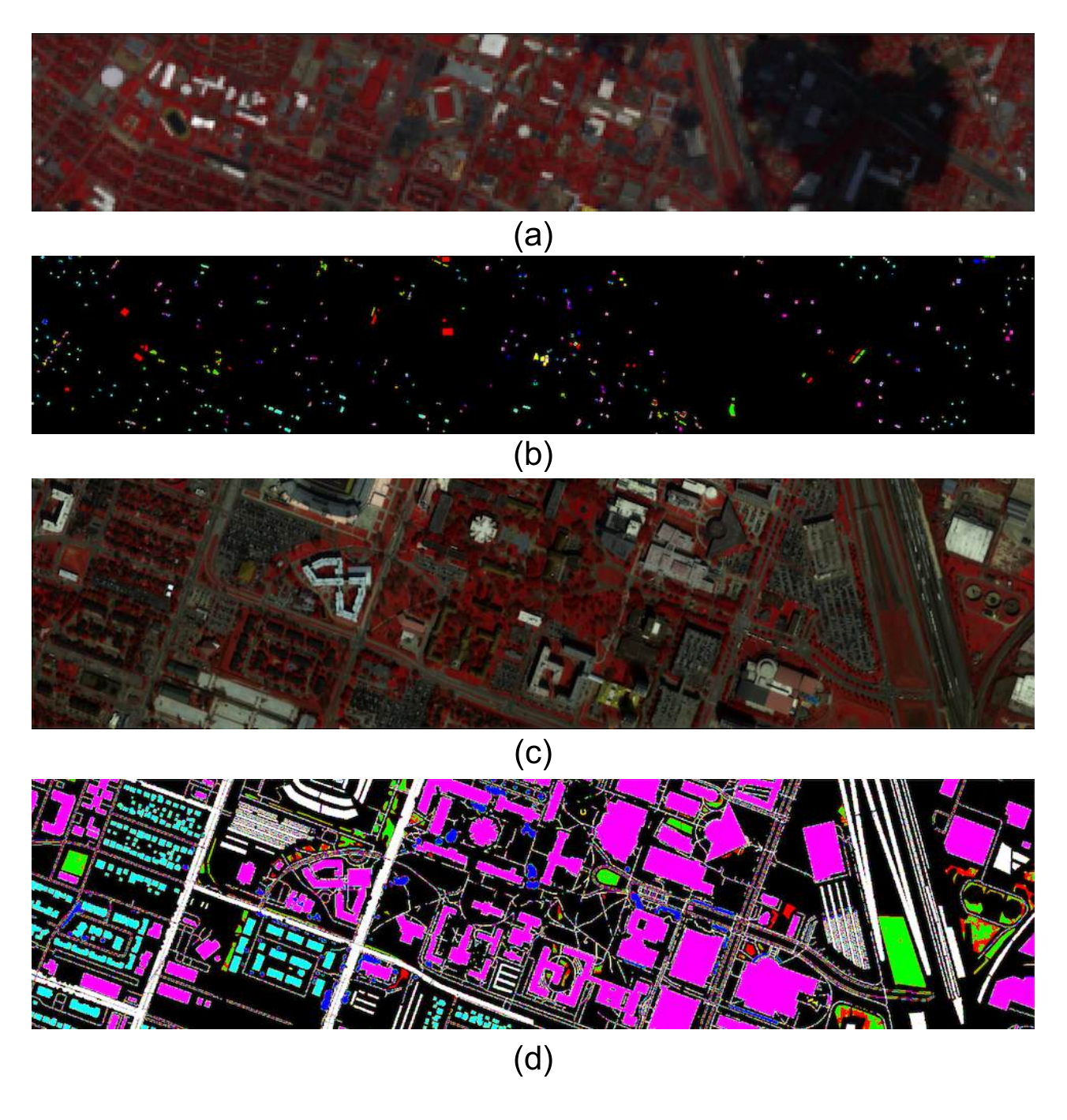}
\caption{HU13–HU18 task. (a) Pseudo-color image of HU13 with 9,176 labeled pixels. (b) Ground-truth image of HU13. (c) Pseudo-color image of HU18 with 504,644 labeled pixels. (d) Ground-truth image of HU18. Class distribution (HU13-HU18): \crule[background]{0.5cm}{0.25cm} Background,
\crule[tree]{0.5cm}{0.25cm} Grass Healthy (1449-9799),
\crule[asphalt]{0.5cm}{0.25cm} Grass Stressed (1444-32502),
\crule[brick]{0.5cm}{0.25cm} Trees (1432-18576),
\crule[bitumen]{0.5cm}{0.25cm} Water (507-266),
\crule[shadow]{0.5cm}{0.25cm} Residential buildings (1464-39794),
\crule[meadow]{0.5cm}{0.25cm} Non-residential buildings (1435-223789),
\crule[baresoil]{0.5cm}{0.25cm} Road (1445-45793),
{\setlength{\fboxsep}{0pt}\fcolorbox{black}{unknown}{\rule{0pt}{0.25cm}\hspace{0.5cm}}} Unknown (0-134125)}
\label{fig:hu13_hu18}
\end{center}
\end{figure}

The Houston dataset contains HU13 (collected by ITERS CASI-1500 in 2013, 349 $\times$ 1905 pixels, 2.5 m resolution, 144 bands) and HU18 (IEEE GRSS Data Fusion 2018 dataset, 601 $\times$ 2384 pixels, 1 m resolution, 380-1050 nm wavelength, 48 bands). Experiments use 48 common bands. Evergreen and deciduous trees in HU18 are merged into trees class, creating seven known classes; remaining 12 classes form unknown class (Fig.~\ref{fig:hu13_hu18}).

\subsubsection{Dioni–Loukia}
\begin{figure}[!t]
\begin{center}
\includegraphics[width=0.7\columnwidth]{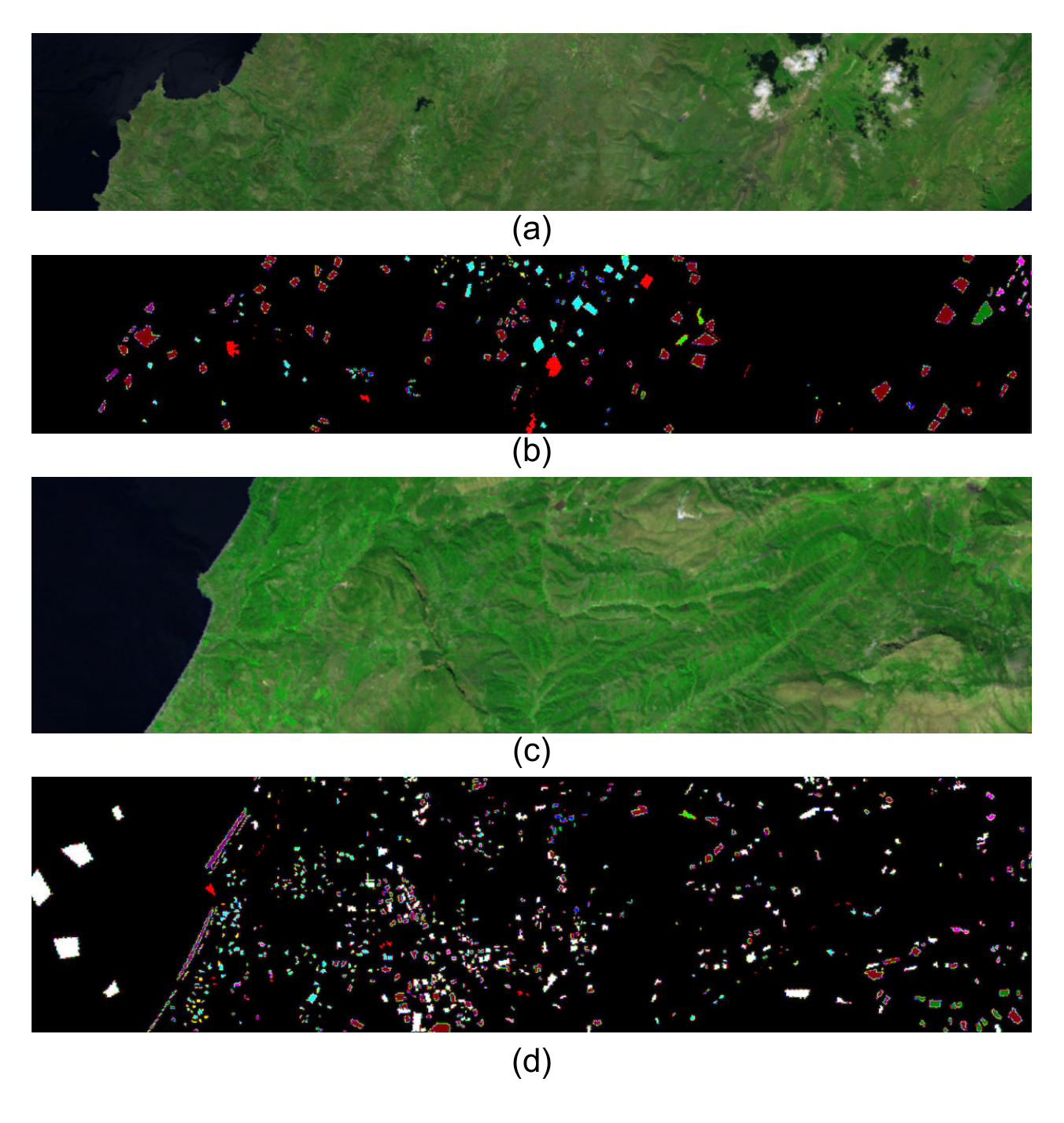}
\caption{Dioni–Loukia task. (a) Pseudo-color image of Dioni with 11,623 labeled pixels. (b) Ground-truth image of Dioni. (c) Pseudo-color image of Loukia with 10,317 labeled pixels. (d) Ground-truth image of Loukia. Class distribution (Dioni-Loukia): \crule[background]{0.5cm}{0.25cm} Background,
\crule[tree]{0.5cm}{0.25cm} Dense Urban Fabric (1262-206),
\crule[asphalt]{0.5cm}{0.25cm} Mineral Extraction Sites (204-54),
\crule[brick]{0.5cm}{0.25cm} Non Irrigated Arable Land (614-426),
\crule[bitumen]{0.5cm}{0.25cm} Fruit Trees (150-79),
\crule[shadow]{0.5cm}{0.25cm} Olive Groves (1768-1107),
\crule[meadow]{0.5cm}{0.25cm} Coniferous Forest (361-422),
\crule[baresoil]{0.5cm}{0.25cm} Sparse Sclerophyllous Vegetation (6374-2361),
\crule[rocksandsand]{0.5cm}{0.25cm} Rocks and Sand (492-453),
\crule[coastalwater]{0.5cm}{0.25cm} Coastal Water (398-421),
{\setlength{\fboxsep}{0pt}\fcolorbox{black}{unknown}{\rule{0pt}{0.25cm}\hspace{0.5cm}}} Unknown (0-4788)}

\label{fig:dioni_loukia}
\end{center}
\end{figure}

Dioni (250 $\times$ 1376 pixels) and Loukia (249 $\times$ 945 pixels) from NASA EO-1 Hyperion sensor, with 176 bands after processing. Nine classes serve as known classes; three classes (dense sclerophyllous vegetation, sparsely vegetated areas, water) in target domain form unknown class (Fig.~\ref{fig:dioni_loukia}).

\subsection{Setup}

\subsubsection{Training details}
All experiments have been conducted on a NVIDIA RTX 3090 GPU. The Adam optimizer \cite{kingma2014adam} was employed with an initial learning rate of $1 \times 10^{-5}$ and weight decay of $1 \times 10^{-5}$. The learning rate was scheduled using CosineAnnealingLR with $T_{\max}$ equal to the total number of epochs and $\eta_{\min} = \text{lr}/10$. Training was conducted for 50 epochs with gradient clipping applied at a maximum norm of 1.0 to ensure training stability.

The input hyperspectral patches were set to $7 \times 7 \times C$ spatial-spectral cubes, where $C$ represents the number of spectral bands (48 for Houston, 102 for Pavia, and 176 for Dioni-Loukia datasets). Batch size was set to 32 for all experiments.

The loss function combines multiple components with dataset-specific optimal weights determined through systematic hyperparameter tuning detailed in Section~\ref{sec:prams}. The regularization weight $\delta = 0.2$ and gradient reversal layer strength $\lambda = 1.0$ were constant across all experiments.

For the training strategy, a single-stage end-to-end approach was adopted where all components were trained jointly. After training completion, threshold calibration using outlier injection was performed targeting 75\% unknown class rejection rate. The calibration employed multiple synthetic unknown generation strategies including Gaussian noise injection, class mixing, spectral corruption, and spatial corruption to optimize uncertainty thresholds (as explained in Section \ref{sec.threshold}).

The best model was selected based on validation accuracy, and final evaluation used calibrated thresholds optimized through ROC analysis on synthetic unknown samples generated during the calibration phase.

\subsubsection{Evaluation Metrics}
Standard open-set recognition metrics were employed \cite{bi2025wgdt, ru2024imbalanced, wan2024unveiling}: Known Class Accuracy (OS) measures classification accuracy on known classes, Unknown Class Rejection Rate (Unk) measures the ability to reject unknown samples, and Harmonic OpenSet Score (HOS) provides a balanced evaluation metric:
\begin{equation}
\text{HOS} = \frac{2 \times \text{OS} \times \text{Unk}}{\text{OS} + \text{Unk}}
\end{equation}
\begin{table*}[t]
\begin{center}
\caption{Classification Results on Three Tasks with the highest value highlighted in red and the second highest in blue.}
\label{tab:combined_results}
\small{
\begin{tabular}{l|ccccccccc|c}
\hline
\multirow{2}{*}{Class} & \multicolumn{9}{c|}{\textbf{Open-set Domain Adaptation}} & \textbf{Open-set DG} \\
\cline{2-11}
& DACD & OSBP & DAMC & STA & MTS & UADAL & OMEGA & ANNA & WGDT& Ours\\
\hline
\multicolumn{11}{c}{\textbf{PU-PC Task}} \\
\hline
1 & 93.2 & \textcolor{blue}{\textbf{95.6}} & 93.3 & 80.0 & 91.0 & 64.1 & 71.3 & 84.6 & \textcolor{red}{\textbf{96.1}} & 87.6 \\
2 & 84.8 & 85.7 & 77.4 & 67.9 & \textcolor{blue}{\textbf{87.6}} & 15.3 & 72.0 & 65.5 & 81.0 & \textcolor{red}{\textbf{99.0}} \\
3 & \textcolor{blue}{\textbf{70.7}} & \textcolor{red}{\textbf{71.7}} & 68.8 & 55.0 & 65.7 & 39.3 & 53.7 & 68.6 & 47.1 & 43.3 \\
4 & 63.1 & \textcolor{red}{\textbf{77.9}} & 63.7 & 54.5 & \textcolor{blue}{\textbf{75.7}} & 45.7 & 25.3 & 65.0 & 47.0 & 52.2 \\
5 & 98.9 & \textcolor{blue}{\textbf{99.2}} & 98.9 & 96.1 & 67.3 & 61.0 & 91.7 & 99.0 & \textcolor{red}{\textbf{100.0}} & 73.9 \\
6 & 48.0 & 39.4 & 29.4 & 21.2 & \textcolor{blue}{\textbf{66.8}} & 55.1 & 65.8 & 64.9 & 64.9 & \textcolor{red}{\textbf{90.8}} \\
7 & \textcolor{blue}{\textbf{44.6}} & 34.3 & 35.2 & 38.4 & 25.8 & 0.8 & 18.3 & 9.2 & \textcolor{red}{\textbf{46.4}} & 9.5 \\
\hline
Unk (\%) & 34.3 & 38.6 & 34.4 & 25.8 & 19.8 & \textcolor{red}{\textbf{98.6}} & 64.9 & 65.9 & 86.3 & \textcolor{blue}{\textbf{91.9}} \\
OS (\%) & \textcolor{blue}{\textbf{71.9}} & \textcolor{red}{\textbf{72.0}} & 66.7 & 63.3 & 68.6 & 40.2 & 56.9 & 65.3 & 68.9 & 66.8 \\
HOS (\%) & 46.3±3.7 & 49.9±6.6 & 45.3±3.1 & 35.4±10.6 & 30.2±7.4 & 56.1±12.6 & 59.2±14.4 & 65.0±4.7 & \textcolor{blue}{\textbf{76.5±4.6}} & \textcolor{red}{\textbf{77.38±3.5}} \\
\hline
\multicolumn{11}{c}{\textbf{HU13-HU18 Task}} \\
\hline
1 & 84.1 & 74.1 & 83.9 & 81.1 & \textcolor{blue}{\textbf{88.3}} & 58.6 & 47.6 & \textcolor{red}{\textbf{94.2}} & 74.5 & 59.1 \\
2 & 34.6 & 28.0 & 31.1 & 25.7 & \textcolor{blue}{\textbf{43.1}} & 33.8 & 37.0 & 41.6 & 34.3 & \textcolor{red}{\textbf{68.9}} \\
3 & 62.5 & 56.2 & 67.8 & \textcolor{red}{\textbf{74.7}} & \textcolor{blue}{\textbf{73.7}} & 36.8 & 49.2 & 51.8 & 48.1 & 61.3 \\
4 & 39.4 & 61.1 & 52.1 & 79.5 & \textcolor{blue}{\textbf{80.8}} & 25.6 & 43.1 & 56.3 & \textcolor{red}{\textbf{90.6}} & 0 \\
5 & 60.5 & 65.4 & 65.9 & \textcolor{red}{\textbf{76.7}} & \textcolor{blue}{\textbf{76.5}} & 19.6 & 35.8 & 42.3 & 65.7 & 28.9 \\
6 & 33.9 & 32.2 & 27.9 & 27.3 & 42.9 & 13.0 & 35.5 & 27.2 & \textcolor{red}{\textbf{64.7}} & \textcolor{blue}{\textbf{61.5}} \\
7 & \textcolor{blue}{\textbf{52.1}} & 51.9 & 47.4 & 40.1 & \textcolor{red}{\textbf{59.4}} & 8.5 & 11.3 & 36.6 & 3.7 & 0 \\
\hline
Unk (\%) & 26.9 & 29.6 & 29.6 & 24.3 & 13.2 & \textcolor{red}{\textbf{89.2}} & 57.8 & 56.0 & \textcolor{blue}{\textbf{70.2}} & 67.3 \\
OS (\%) & 52.4 & 52.7 & 54.0 & \textcolor{blue}{\textbf{57.7}} & \textcolor{red}{\textbf{66.4}} & 28.0 & 37.1 & 50.0 & 54.5 & 50.9 \\
HOS (\%) & 35.0±5.4 & 37.6±3.7 & 37.7±4.9 & 33.1±8.2 & 21.4±8.1 & 41.5±12.5 & 43.7±9.2 & 51.7±5.6 & \textcolor{red}{\textbf{61.1±2.8}} & \textcolor{blue}{\textbf{57.97±2.4}} \\
\hline
\multicolumn{11}{c}{\textbf{Dioni-Loukia Task}} \\
\hline
1 & 12.9 & \textcolor{red}{\textbf{24.8}} & \textcolor{blue}{\textbf{19.3}} & 9.5 & 16.9 & 4.9 & 10.8 & 8.5 & 3.9 & 0 \\
2 & 48.1 & 43.5 & 57.2 & 68.7 & 83.9 & \textcolor{blue}{\textbf{96.7}} & \textcolor{red}{\textbf{97.2}} & 87.6 & 88.0 & 92.6 \\
3 & 21.7 & 33.4 & 31.3 & 41.5 & 28.5 & 28.0 & 19.3 & 21.7 & \textcolor{red}{\textbf{51.5}} & \textcolor{blue}{\textbf{42.0}} \\
4 & 70.4 & 72.0 & 70.4 & \textcolor{blue}{\textbf{76.5}} & \textcolor{red}{\textbf{84.1}} & 55.7 & 39.4 & 49.4 & 56.3 & 35.4 \\
5 & 8.0 & 13.1 & 13.2 & \textcolor{red}{\textbf{22.3}} & 7.9 & 2.5 & \textcolor{blue}{\textbf{17.6}} & 2.9 & 0.1 & 0 \\
6 & 62.8 & 58.9 & 56.9 & 39.8 & \textcolor{red}{\textbf{73.3}} & 36.4 & 38.1 & 57.7 & 63.7 & \textcolor{blue}{\textbf{64.9}} \\
7 & 11.6 & \textcolor{blue}{\textbf{17.1}} & 11.7 & 3.5 & 11.1 & 3.0 & 2.8 & 11.8 & 5.7 & \textcolor{red}{\textbf{47.2}} \\
8 & 3.1 & 2.0 & 4.7 & 7.4 & 24.7 & \textcolor{blue}{\textbf{38.1}} & 10.9 & 10.3 & 52.7 & \textcolor{red}{\textbf{74.0}} \\
9 & 28.5 & 8.7 & 18.0 & 61.8 & 82.3 & \textcolor{blue}{\textbf{85.7}} & 83.6 & 14.8 & \textcolor{red}{\textbf{97.7}} & 54.9 \\
\hline
Unk (\%) & 23.4 & 25.0 & 23.0 & 13.9 & 6.2 & 48.8 & \textcolor{blue}{\textbf{54.0}} & 46.4 & 52.0 & \textcolor{red}{\textbf{56.02}} \\
OS (\%) & 29.7 & 30.4 & 31.4 & 36.8 & \textcolor{blue}{\textbf{45.6}} & 39.0 & 35.5 & 29.4 & \textcolor{red}{\textbf{46.6}} & 39.99 \\
HOS (\%) & 25.3±5.2 & 26.4±3.1 & 25.7±5.3 & 18.5±10.2 & 10.3±7.0 & 41.6±4.3 & 40.4±10.3 & 34.0±6.1 & \textcolor{red}{\textbf{48.9±1.3}} & \textcolor{blue}{\textbf{46.66±3.2}} \\
\hline
\end{tabular}}
\end{center}
\end{table*}
The HOS score provides a single comprehensive metric that balances the trade-off between maintaining high accuracy on known classes while effectively detecting unknown classes.


All experiments were conducted 5 times with different random seeds to ensure statistical significance; the mean results are reported.

\subsection{Comparison against existing approaches}
Since there are no reported methods for OSDG in hyperspectral image classification, The proposed method has been compared against OSDA models: DACD \cite{zhao2022dual}, OSBP \cite{saito2018open}, DAMC \cite{shermin2021adversarial}, STA \cite{liu2019separate}, MTS \cite{chang2024mind}, UADAL \cite{jang2022unknown}, OMEGA \cite{ru2024imbalanced}, ANNA \cite{li2023adjustment}, and the recent hyperspectral-specific method WGDT \cite{bi2025wgdt}.

\begin{figure*}[t]
  \begin{center}
  \subfloat[Ground Truth]{\label{fig.pavia.f}\includegraphics[width=0.149\textwidth]{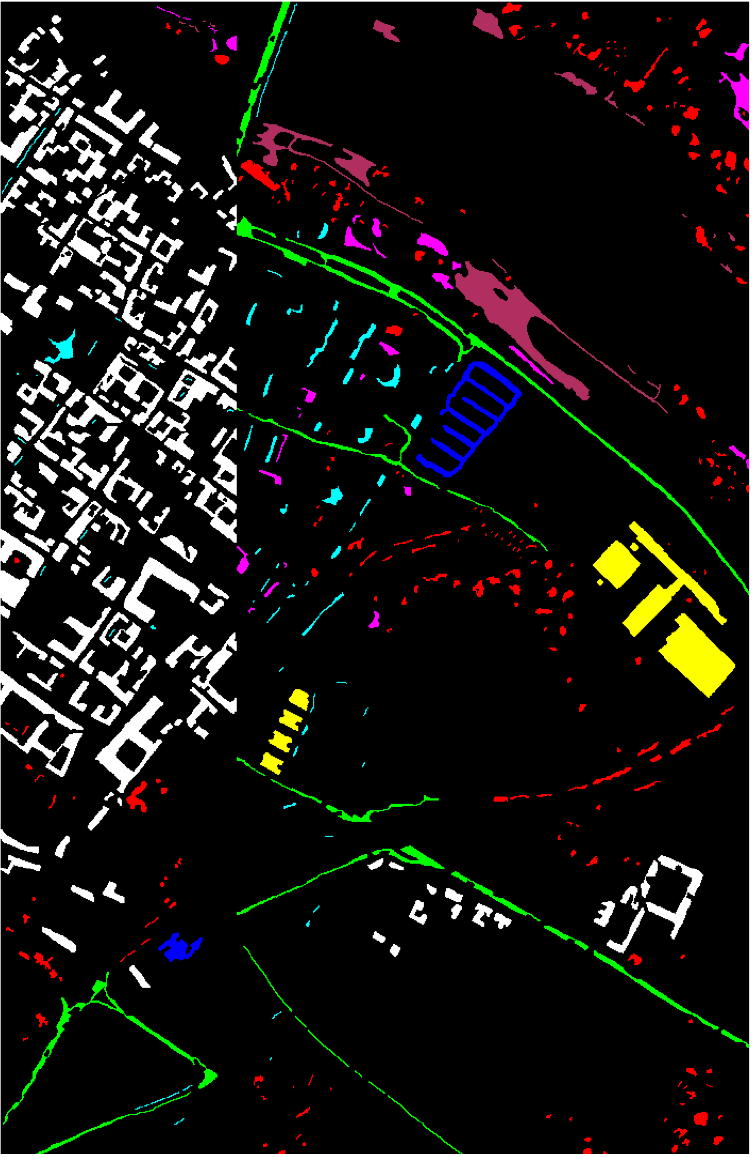}}\,
  \subfloat[OMEGA]{\label{fig.pavia.g}\includegraphics[width=0.15\textwidth]{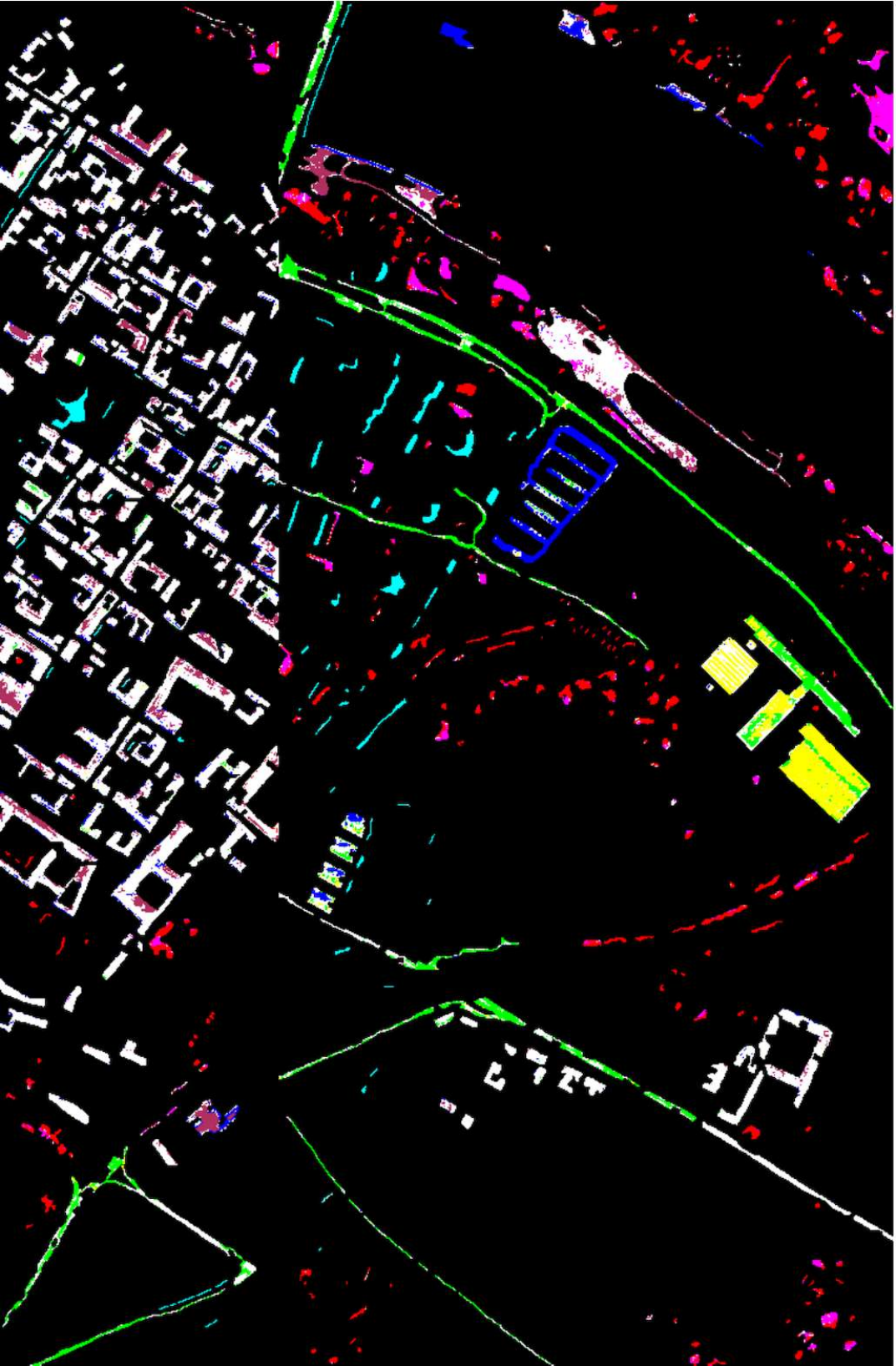}}\,
  \subfloat[ANNA]{\label{fig.pavia.h}\includegraphics[width=0.15\textwidth]{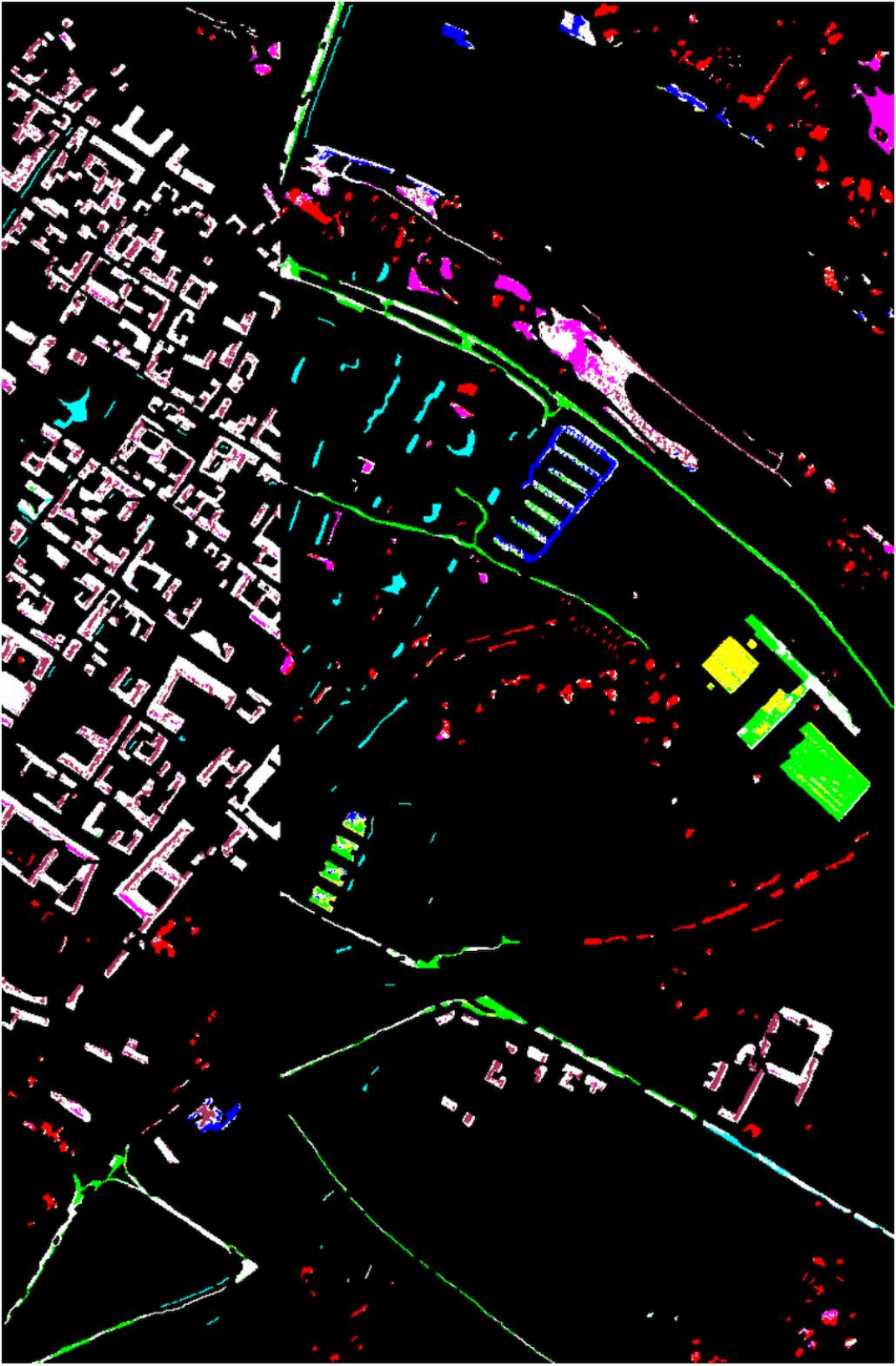}}\,
  \subfloat[WGDT]{\label{fig.pavia.i}\includegraphics[width=0.15\textwidth]{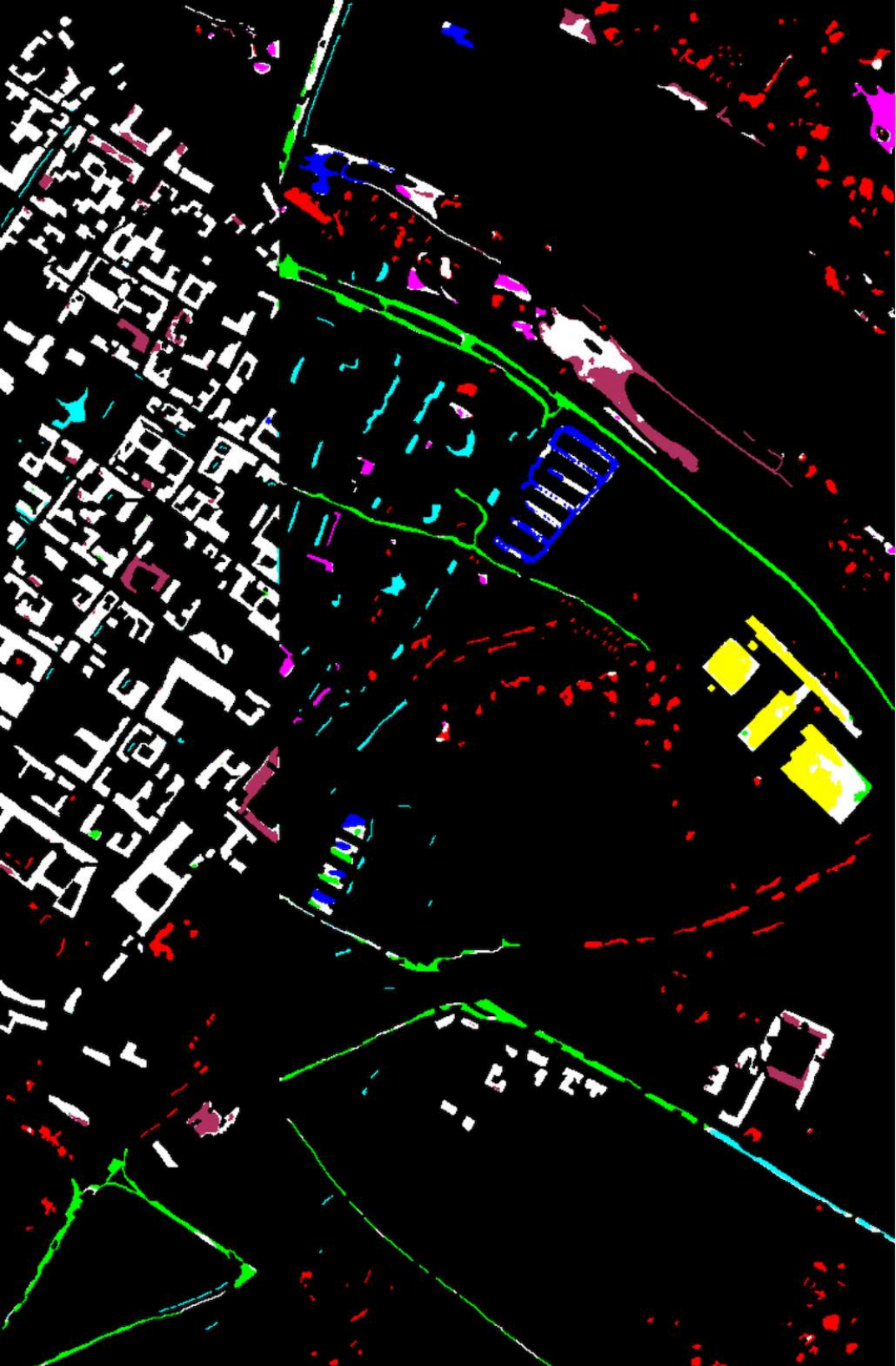}}\,
  \subfloat[Ours]{\label{fig.pavia.j}\includegraphics[width=0.15\textwidth]{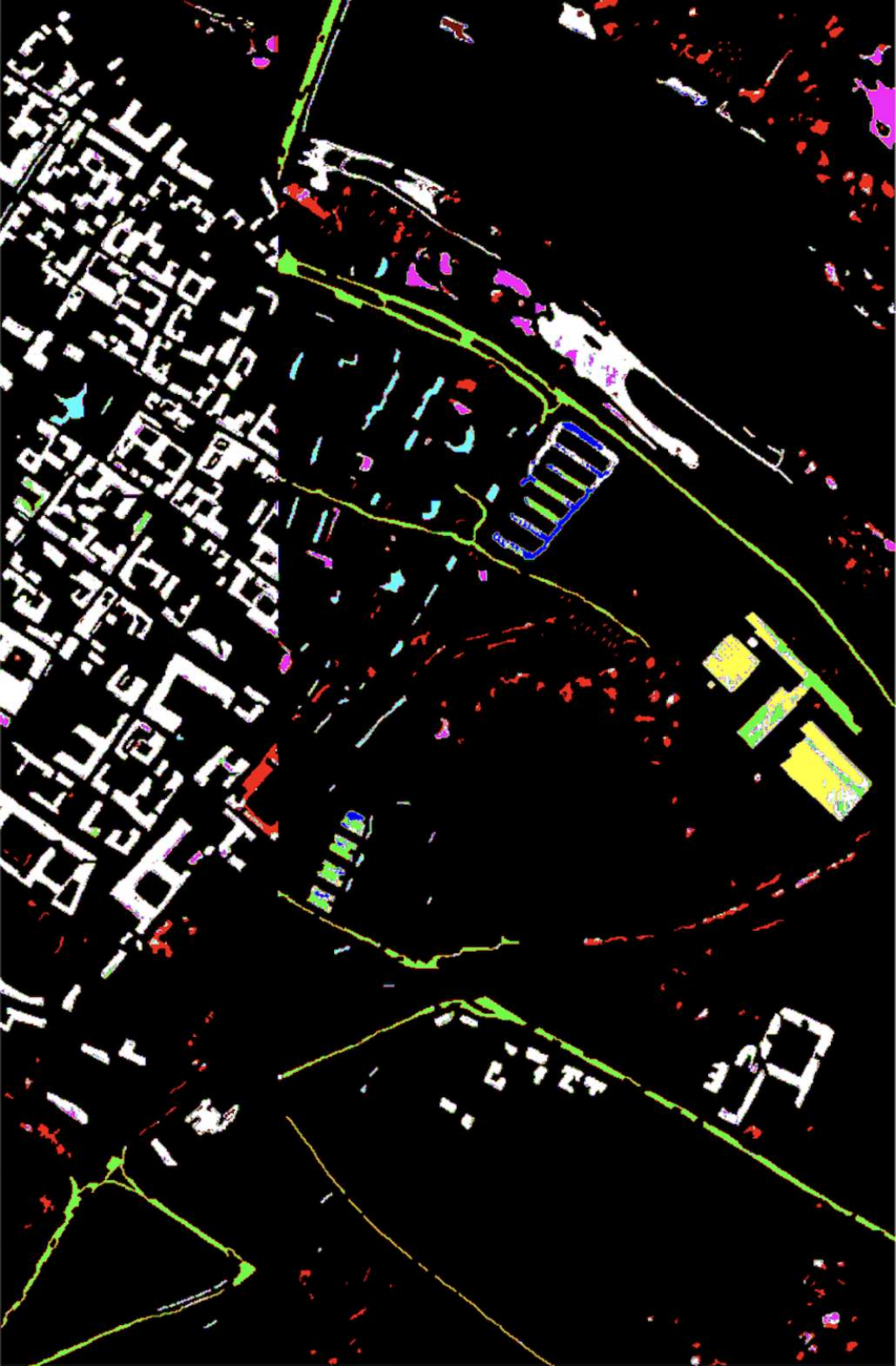}}
  \caption{Classification map of the PU-PC task for the ground truth and the top 4 performing models. \crule[background]{0.5cm}{0.25cm} Background,
\crule[brick]{0.5cm}{0.25cm} Brick,
\crule[meadow]{0.5cm}{0.25cm} Meadow,
\crule[tree]{0.5cm}{0.25cm} Tree,
\crule[bitumen]{0.5cm}{0.25cm} Bitumen,
\crule[baresoil]{0.5cm}{0.25cm} Bare Soil,
\crule[asphalt]{0.5cm}{0.25cm} Asphalt,
\crule[shadow]{0.5cm}{0.25cm} Shadow,
{\setlength{\fboxsep}{0pt}\fcolorbox{black}{unknown}{\rule{0pt}{0.25cm}\hspace{0.5cm}}} Unknown}
  \label{fig:pavia-classification}
\end{center}
\end{figure*}

\begin{figure*}[t]
  \begin{center}
  \subfloat[Ground Truth]{\label{fig.dioni.g}\includegraphics[width=0.483\textwidth]{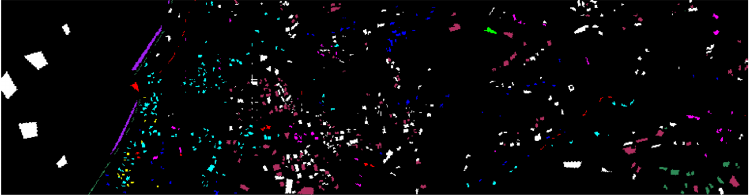}}\,
  \subfloat[UADL]{\label{fig.dioni.h}\includegraphics[width=0.48\textwidth]{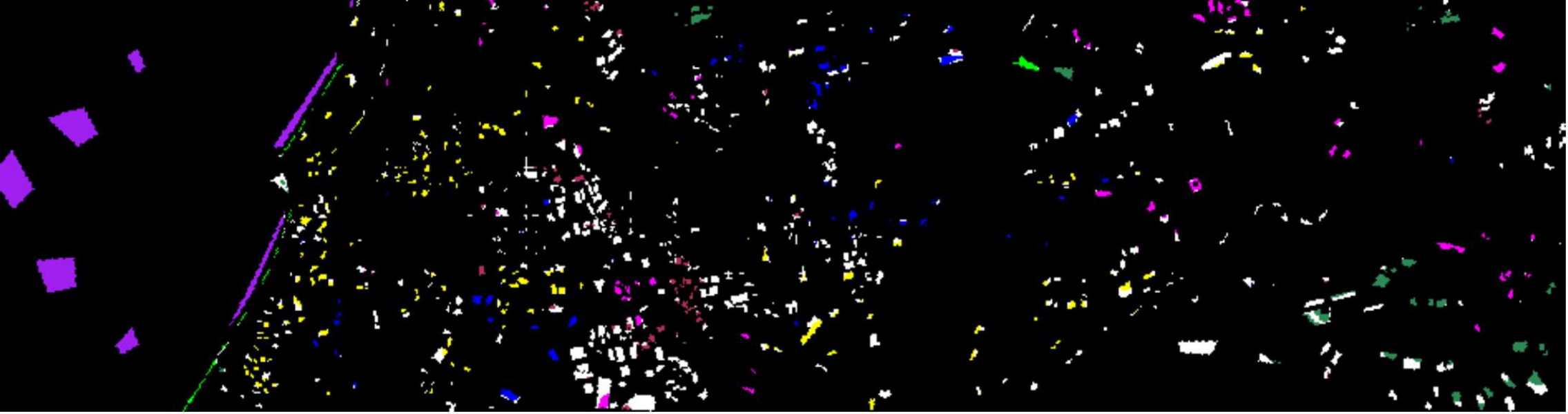}}\\
  \subfloat[WGDT]{\label{fig.dioni.i}\includegraphics[width=0.48\textwidth]{dioni/i.pdf}}\,
  \subfloat[Ours]{\label{fig.dioni.j}\includegraphics[width=0.48\textwidth]{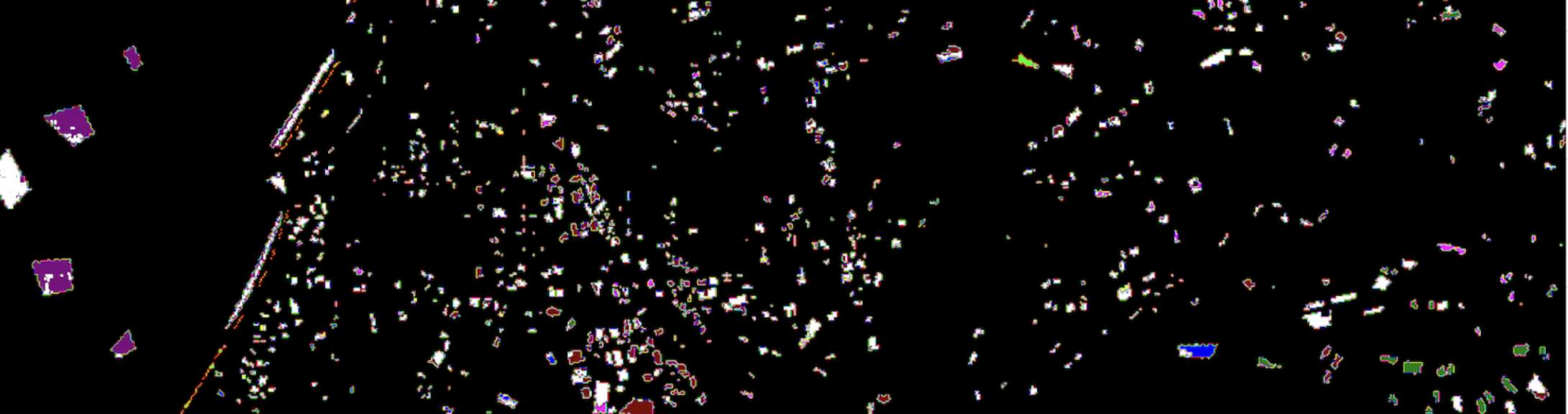}}
  \caption{Classification map of the Dioni–Louki task for the ground truth and the top 3 performing models. \crule[background]{0.5cm}{0.25cm} Background,
\crule[tree]{0.5cm}{0.25cm} Dense Urban Fabric,
\crule[asphalt]{0.5cm}{0.25cm} Mineral Extraction Sites,
\crule[brick]{0.5cm}{0.25cm} Non Irrigated Arable Land,
\crule[bitumen]{0.5cm}{0.25cm} Fruit Trees,
\crule[shadow]{0.5cm}{0.25cm} Olive Groves,
\crule[meadow]{0.5cm}{0.25cm} Coniferous Forest,
\crule[baresoil]{0.5cm}{0.25cm} Sparse Sclerophyllous Vegetation,
\crule[rocksandsand]{0.5cm}{0.25cm} Rocks and Sand,
\crule[coastalwater]{0.5cm}{0.25cm} Coastal Water,
{\setlength{\fboxsep}{0pt}\fcolorbox{black}{unknown}{\rule{0pt}{0.25cm}\hspace{0.5cm}}} Unknown}
  \label{fig:dioni-classification}
\end{center}
\end{figure*}

\begin{figure*}[t]
  \begin{center}
  \subfloat[Ground Truth]{\label{fig.houston.g}\includegraphics[width=0.49\textwidth]{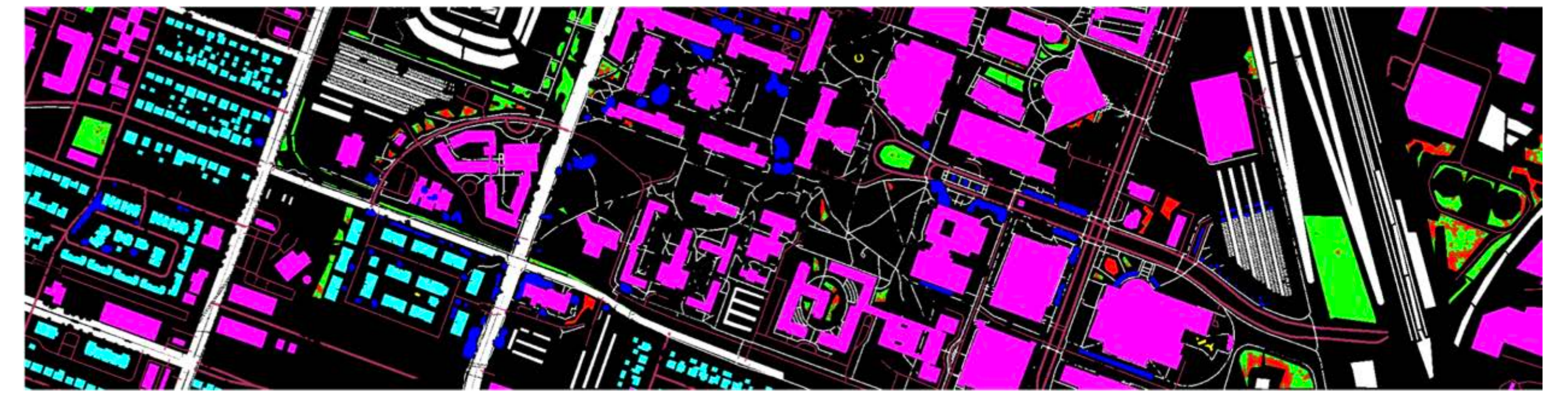}}\,
  \subfloat[ANNA]{\label{fig.houston.h}\includegraphics[width=0.48\textwidth]{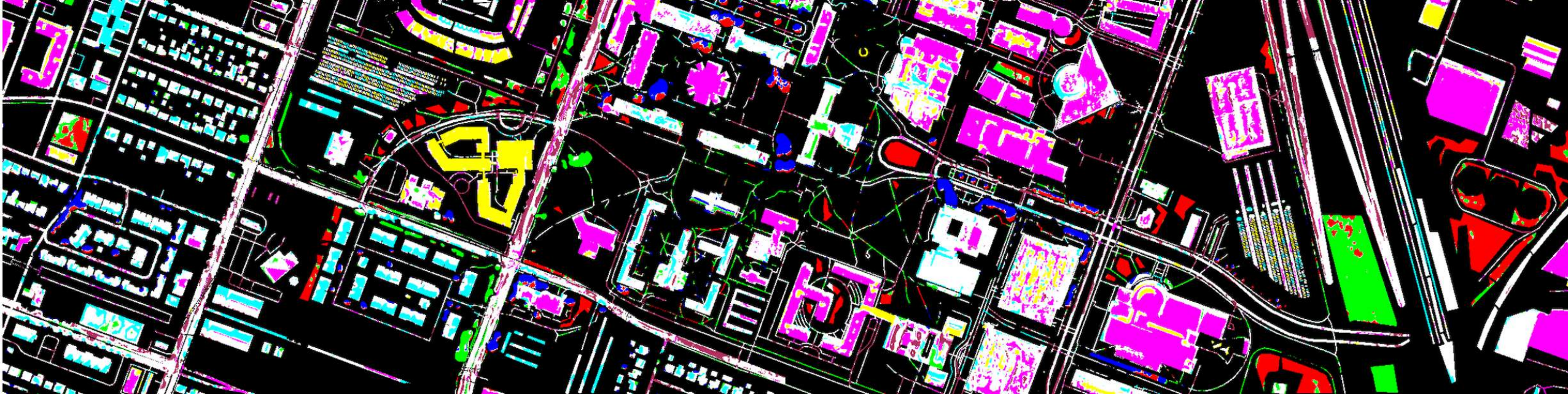}}\\
  \subfloat[WGDT]{\label{fig.houston.i}\includegraphics[width=0.48\textwidth]{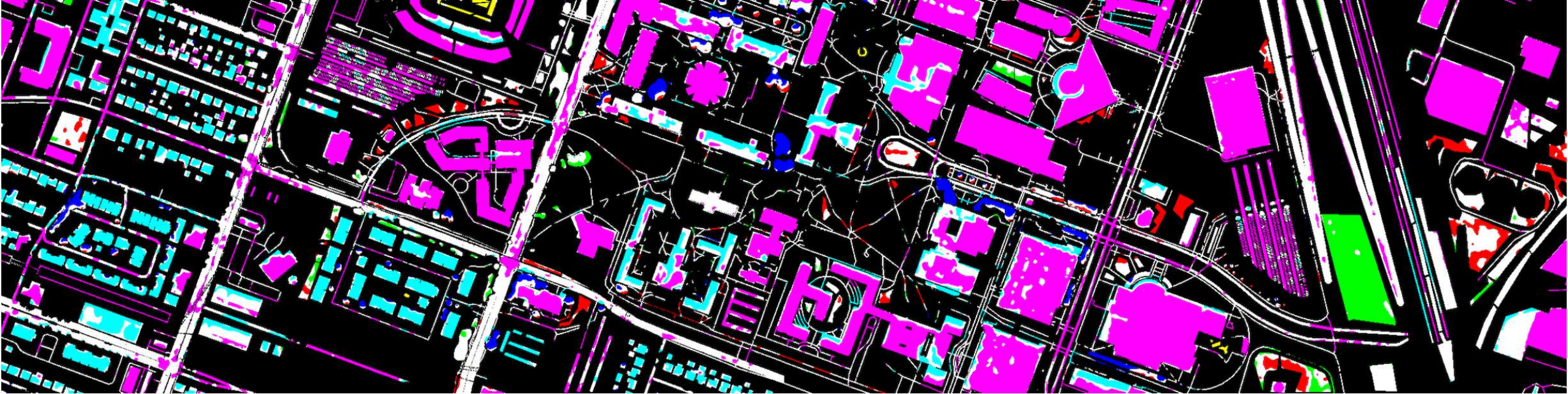}}\,
  \subfloat[Ours]{\label{fig.houston.j}\includegraphics[width=0.48\textwidth]{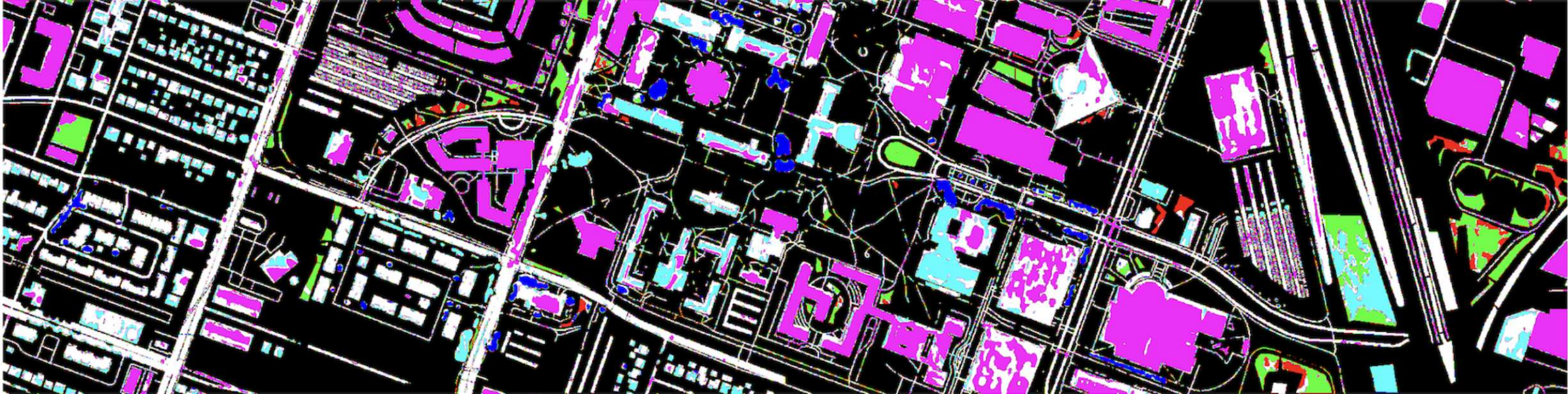}}
  \caption{Classification map of the HU13–HU18 task for the ground truth and the top 3 performing models. \crule[background]{0.5cm}{0.25cm} Background,
\crule[tree]{0.5cm}{0.25cm} Grass Healthy,
\crule[asphalt]{0.5cm}{0.25cm} Grass Stressed,
\crule[brick]{0.5cm}{0.25cm} Trees,
\crule[bitumen]{0.5cm}{0.25cm} Water,
\crule[shadow]{0.5cm}{0.25cm} Residential buildings,
\crule[meadow]{0.5cm}{0.25cm} Non-residential buildings,
\crule[baresoil]{0.5cm}{0.25cm} Road,
{\setlength{\fboxsep}{0pt}\fcolorbox{black}{unknown}{\rule{0pt}{0.25cm}\hspace{0.5cm}}} Unknown}
  \label{fig:houston-classification}
\end{center}
\end{figure*}
\subsection{Results and Discussion}


The results (Table \ref{tab:combined_results}) reveal several key insights about the proposed approach. On the PU-PC task, this method achieves the highest HOS score, significantly outperforming all baseline methods. This superior performance is primarily attributed to the proposed method's excellent unknown class rejection capabilities, which show the effectiveness of the spectral-spatial uncertainty disentanglement mechanism. While the known class accuracy remains competitive, the balanced performance between known and unknown class detection results in the optimal harmonic mean. For the HU13-HU18 task, the approach ranks second after WGDT. This task presents particular challenges due to the significant domain shift between the 2013 and 2018 Houston datasets, which differ in spatial resolution and spectral characteristics. The proposed method shows robust unknown class detection while maintaining reasonable known class performance, indicating effective generalization across temporal variations. The Dioni-Loukia task shows the proposed method ranking second after WGDT. This result is particularly noteworthy given that the proposed method operates under the DG paradigm \textbf{without accessing target domain data during training}, and still achieves performance comparable to sophisticated DA methods.

However, the proposed method exhibits complete classification failure on four specific classes: Water and Road in HU13-HU18, and Dense Urban Fabric and Olive Groves in Dioni-Loukia. This failure pattern suggests that the uncertainty-driven rejection mechanism in SSUD may be overly conservative for certain spectral signatures, systematically rejecting these classes as unknown rather than classifying them correctly. Additionally, the domain-invariant frequency processing in SIFD may inadvertently filter out discriminative spectral features crucial for these land cover types, particularly those with high spectral variability or domain-dependent characteristics. These failures highlight a critical trade-off in OSDG: while the framework successfully maintains high unknown class rejection rates, it sacrifices performance on challenging known classes that exhibit domain-dependent spectral signatures. This limitation is particularly evident in spectrally variable classes like water bodies, affected by atmospheric conditions and depth variations, and urban areas, with varying material compositions across geographical locations, suggesting that future improvements should investigate class-specific uncertainty thresholds to better balance known class accuracy with unknown class detection.

\subsection{Uncertainty and Evidence Analysis}
To provide deeper insights into the uncertainty quantification mechanisms and validate the effectiveness of the SSUD framework, an uncertainty analysis has been conducted on the PC dataset. 

\begin{figure}[t]
  \begin{center}
  \includegraphics[width=0.7\columnwidth]{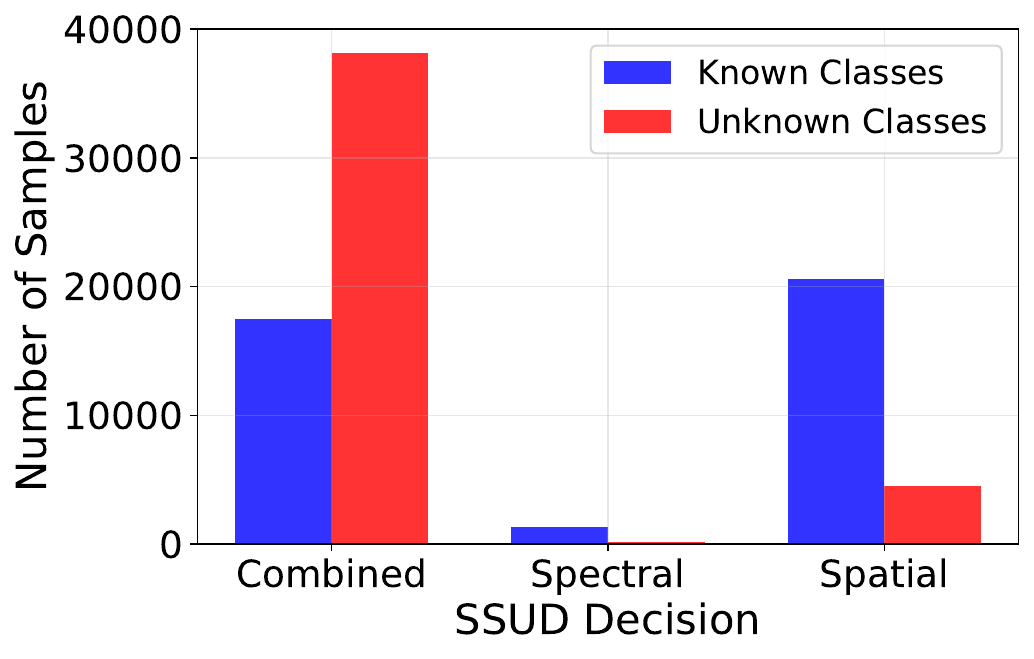}
  \caption{SSUD decision distribution showing pathway selection preferences for known and unknown classes on the PC dataset.}
  \label{fig:ssud_decisions}
  \end{center}
\end{figure}

\textbf{Decision Distribution Analysis:} The pathway selection statistics (Fig.~\ref{fig:ssud_decisions}) show that the SSUD adapts appropriately to class characteristics. For known classes, it predominantly relies on the combined pathway and spatial pathway, while spectral decisions are less frequent. Unknown classes show a different pattern with higher reliance on combined decisions, indicating that uncertainty-driven pathway selection effectively handles novel spectral signatures.

\begin{figure}[t]
  \begin{center}
  \includegraphics[width=0.7\columnwidth]{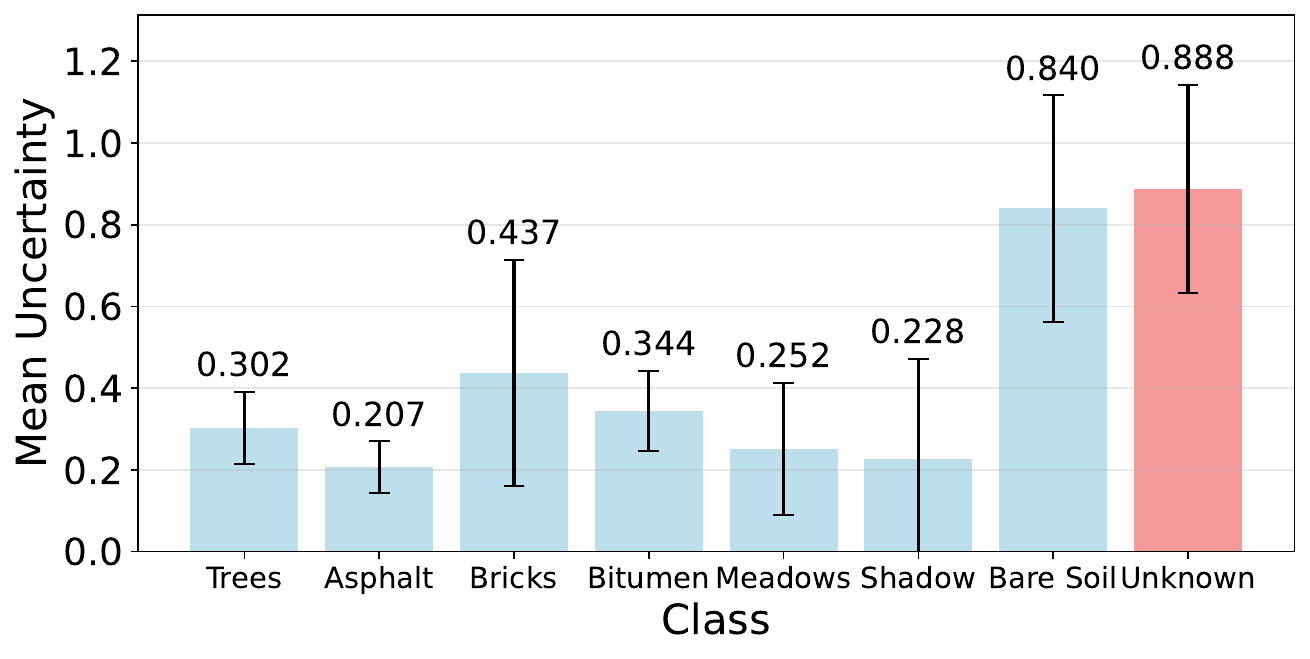}
  \caption{Class-wise uncertainty patterns for the PC dataset, showing mean uncertainty values with standard deviations for each land cover class.}
  \label{fig:class_uncertainty}
  \end{center}
\end{figure}

\textbf{Class-Wise Uncertainty Patterns:} Uncertainty quantification (Fig.~\ref{fig:class_uncertainty}) reveals meaningful patterns across different land cover types. Natural classes like Trees and Asphalt exhibit lower uncertainties, reflecting their consistent spectral characteristics. In contrast, spectrally variable classes like Bricks show higher uncertainty. Most importantly, the Unknown class achieves the highest uncertainty, demonstrating effective separation from known classes.

\begin{figure}[t]
  \begin{center}
  \includegraphics[width=0.7\columnwidth]{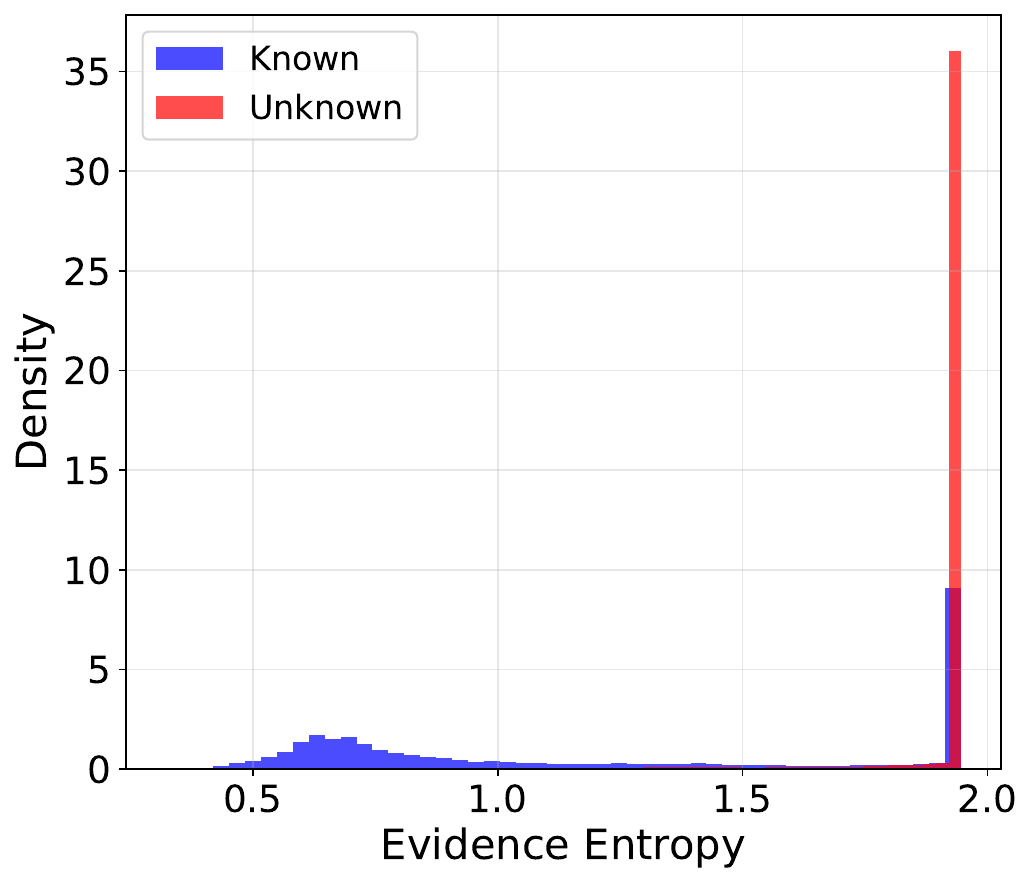}
  \caption{Evidence concentration analysis showing the distribution of evidence entropy for known and unknown classes on the PC dataset.}
  \label{fig:evidence_concentration}
  \end{center}
\end{figure}

\textbf{Evidence Concentration Analysis:} Evidence entropy distribution (Figu.~\ref{fig:evidence_concentration}) distinguishes between known and unknown classes. Known classes exhibit concentrated evidence with relatively low entropy values, indicating confident predictions with concentrated Dirichlet distributions. In contrast, unknown classes show dispersed evidence with significantly higher entropy, reflecting the uniform distribution characteristic of uncertain predictions. This separation validates the evidential learning approach for OSDG.

\begin{figure}[t]
  \begin{center}
  \includegraphics[width=0.7\columnwidth]{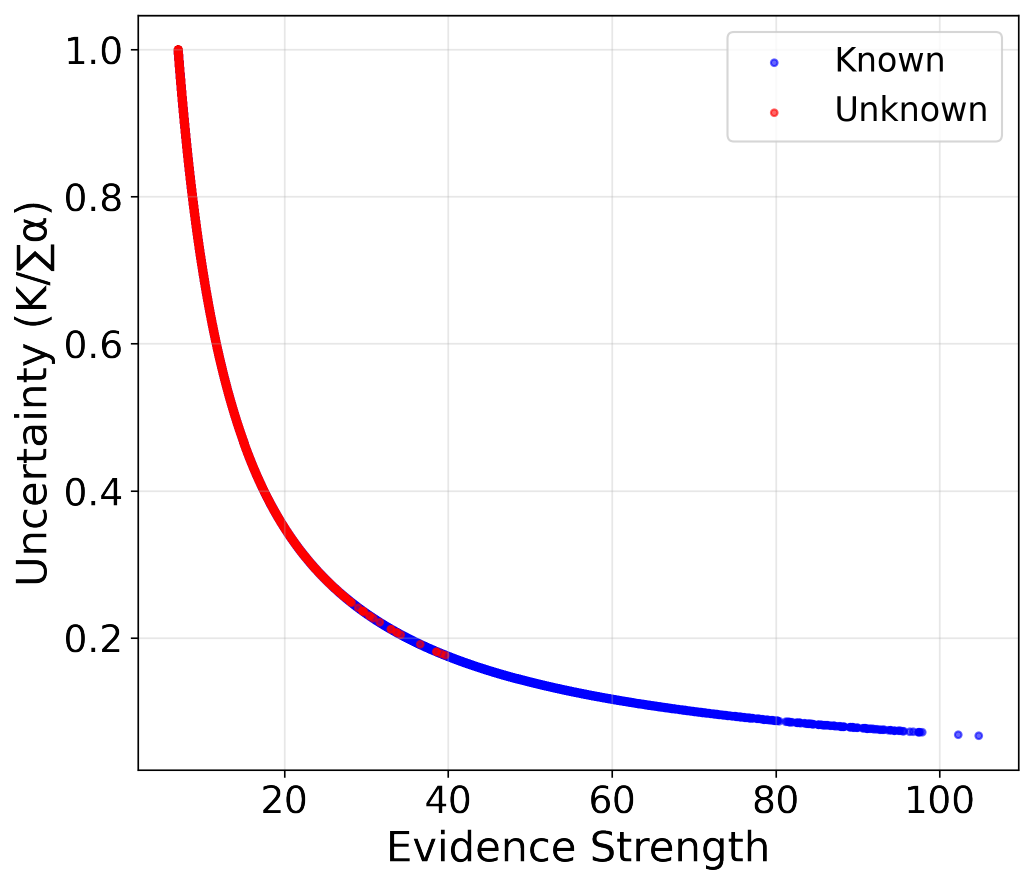}
  \caption{Evidence vs uncertainty relationship demonstrating the inverse correlation between evidence strength and uncertainty for both known and unknown classes on the PC dataset.}
  \label{fig:evidence_uncertainty}
  \end{center}
\end{figure}

\textbf{Evidence-Uncertainty Relationship:} The inverse relationship between evidence strength and uncertainty (Fig.~\ref{fig:evidence_uncertainty}) follows theoretical expectations, with both known (blue) and unknown (red) classes exhibiting decreasing uncertainty as evidence increases. However, unknown classes consistently maintain higher uncertainty levels across all evidence strengths, providing a reliable mechanism for rejection decisions. This relationship validates the evidential framework's ability to provide meaningful uncertainty estimates that correlate with prediction confidence.

\begin{figure}[t]
  \begin{center}
  \includegraphics[width=0.7\columnwidth]{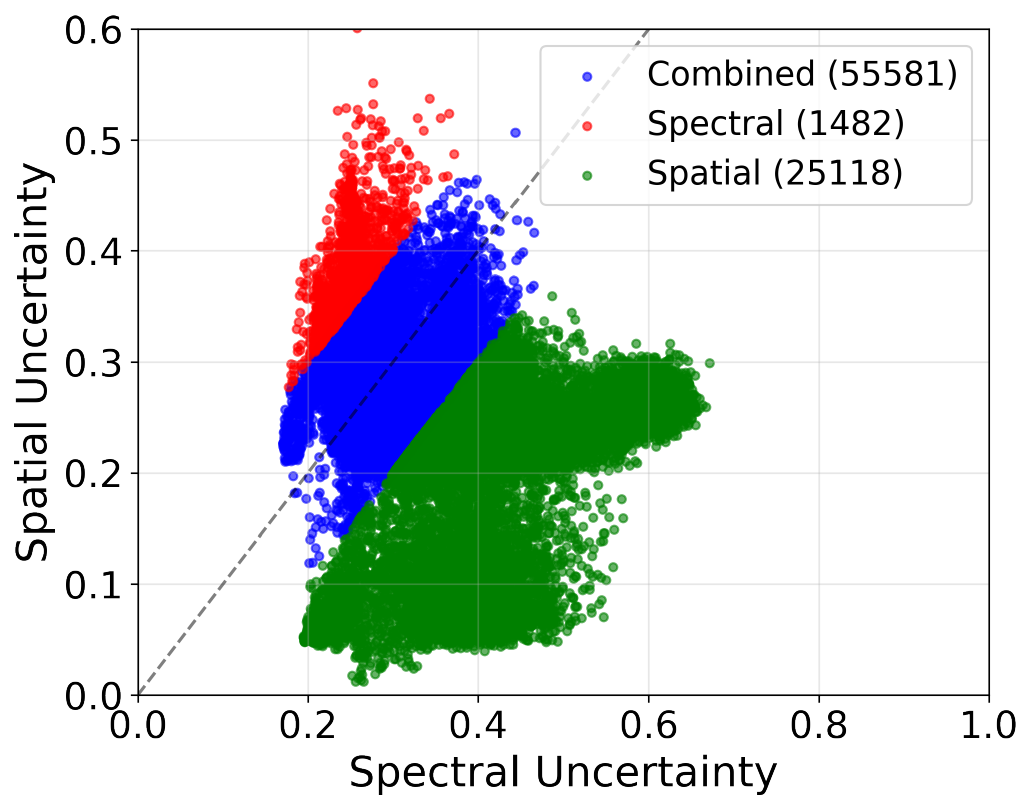}
  \caption{Individual pathway uncertainties with SSUD decisions, showing the relationship between spectral and spatial uncertainties and the corresponding pathway selection strategy on the PC dataset.}
  \label{fig:pathway_uncertainties}
  \end{center}
\end{figure}

\textbf{Pathway-Specific Uncertainty Relationships:} Figure \ref{fig:pathway_uncertainties} reveals how SSUD decides based on the relationship between spectral and spatial pathway uncertainties. Points below the diagonal line indicate scenarios where spectral uncertainty dominates, leading to spatial pathway selection (green points), while points above favor spectral pathway decisions (red points). The combined pathway (blue points) is selected when uncertainties are balanced. This visualization validates the theoretical assumption that different pathways exhibit varying reliability across samples, with the SSUD mechanism effectively leveraging this complementary information.

These visualizations collectively show that the uncertainty quantification framework provides interpretable and reliable estimates that effectively support OSDG decisions.

\subsection{Ablation study}
To validate the effectiveness of each module, comprehensive ablation studies have been conducted on the PU - PC datasets.

\textbf{SIFD:}
Eight ablation experiments were designed to isolate the contribution of each SIFD component (Table \ref{tab:sifd_ablation}). In \textit{NoFreqTransform}, frequency domain processing is removed entirely. In \textit{NoAttention}, the attention mechanism that weights frequency components is eliminated, with uniform weighting used instead. In \textit{NoDomainReg}, the domain adversarial regularization term is removed. In \textit{NoRecon}, the spectrum reconstruction loss is removed. For frequency transform alternatives, in \textit{DCT}, FFT is replaced with DCT; in \textit{Wavelet}, FFT is substituted with a wavelet transform for multi-resolution analysis; in \textit{RealOnly}, only the real components of the FFT output are used; and in \textit{ImagOnly}, only the imaginary components are used to assess the relative importance of magnitude versus phase information.

\begin{table}[h]
\begin{center}
\caption{SIFD Ablation Study Results on PU - PC Task}
\label{tab:sifd_ablation}
\small{
\begin{tabular}{lccc}
\toprule
\textbf{Variant} & \textbf{OS* (\%)} & \textbf{Unk (\%)} & \textbf{HOS (\%)} \\
\midrule
Original & 66.80 & \textcolor{red}{\textbf{91.90}} & \textcolor{red}{\textbf{77.38}} \\
RealOnly & 65.17 & \textcolor{blue}{\textbf{84.39}} & \textcolor{blue}{\textbf{73.54}} \\
DCT & \textcolor{blue}{\textbf{72.33}} & 53.74 & 61.66 \\
NoRecon & 69.02 & 40.82 & 51.30 \\
Wavelet & 68.03 & 33.62 & 45.00 \\
NoFreqTransform & 62.93 & 28.25 & 38.99 \\
ImagOnly & 66.59 & 22.53 & 33.67 \\
NoAttention & \textcolor{red}{\textbf{75.25}} & 19.48 & 30.94 \\
NoDomainReg & 70.64 & 16.60 & 26.88 \\
\bottomrule
\end{tabular}}
\end{center}
\end{table}

\textit{RealOnly} achieves competitive performance, demonstrating that real FFT components contain most domain-invariant information, while \textit{ImagOnly} performs poorly, indicating imaginary components primarily capture domain-specific phase variations. \textit{DCT} achieves high known class accuracy but poor unknown rejection, suggesting it preserves discriminative features but lacks uncertainty estimation properties. Removing core components severely degrades performance: \textit{NoFreqTransform} confirms frequency analysis is essential, \textit{NoAttention} produces dramatic drops in unknown rejection despite high known accuracy, revealing attention-weighted selection is crucial for domain-invariant features. \textit{NoDomainReg} shows catastrophic failure, confirming domain adversarial training prevents overfitting to source characteristics, while \textit{NoRecon} shows reconstruction loss serves as crucial regularization. Thus, the original SIFD design achieves optimal synthesis, validating that each component contributes meaningfully to the framework's effectiveness.

\textbf{DCRN:}
Six ablation experiments were designed to isolate the contribution of each DCRN component (Table \ref{tab:dcrn_ablation}). In \textit{SpectralOnly}, the spatial processing pathway is removed. In \textit{ElementwiseAddition}, the attention-based fusion mechanism is replaced with simple element-wise addition, while in \textit{SimpleConcatenation}, attention fusion is substituted with basic concatenation. For backbone architecture alternatives, in \textit{ResNet50}, the entire DCRN is replaced with a standard ResNet50 architecture; in \textit{VGG}, DCRN is substituted with VGG-style layers; and in \textit{EfficientNet}, DCRN is replaced with EfficientNet to compare the dual-pathway design against established convnets.

\begin{table}[h]
\begin{center}
\caption{DCRN Ablation Study Results on PU - PC Task}
\label{tab:dcrn_ablation}
\small{
\begin{tabular}{lccc}
\toprule
\textbf{Variant} & \textbf{OS* (\%)} & \textbf{Unk (\%)} & \textbf{HOS (\%)} \\
\midrule
Original & \textcolor{red}{\textbf{66.80}} & \textcolor{red}{\textbf{91.90}} & \textcolor{red}{\textbf{77.38}} \\
SpectralOnly & \textcolor{blue}{\textbf{66.67}} & \textcolor{blue}{\textbf{91.72}} & \textcolor{blue}{\textbf{77.22}} \\
ElementwiseAddition & 66.64 & 91.68 & 77.18 \\
SimpleConcatenation & 66.38 & 91.32 & 76.89 \\
ResNet50 & 63.79 & 87.76 & 73.87 \\
VGG & 62.94 & 86.59 & 72.87 \\
EfficientNet & 60.23 & 82.87 & 69.74 \\
\bottomrule
\end{tabular}}
\end{center}
\end{table}

The results reveal that spectral processing dominates performance, with \textit{SpectralOnly} achieving near-optimal results, showing that spectral pathways contain most domain-invariant information. Fusion mechanism choices prove less critical, as \textit{ElementwiseAddition} and \textit{SimpleConcatenation} perform nearly identically to the original, indicating the dual-pathway architecture's strength lies in complementary feature extraction rather than sophisticated fusion strategies. Standard convnets show notable performance degradation compared to the original DCRN design. ResNet50, VGG, and EfficientNet all show declining HOS scores, with EfficientNet showing the most significant drop. This confirms that HSI requires domain-specific architectural considerations rather than generic optimization, confirming that the original DCRN design achieves optimal balance through its tailored dual-pathway.

Notably, the consistent performance across different fusion mechanisms and the relatively stable results when varying architectural components show the robustness and generalizability of the proposed method. This architectural flexibility suggests that it can be readily adapted to different computational constraints and implementation requirements while maintaining strong OSDG performance, rendering it applicable across diverse HSI scenarios.

\textbf{EDL:}
Five ablation experiments were designed to replace EDL with alternative uncertainty quantification methods (Table \ref{tab:edl_ablation}). \textit{TempScaling} uses learnable temperature parameters for calibrated softmax probabilities. \textit{Entropy} employs normalized entropy $-\sum p_i \log p_i / \log(K)$ as uncertainty. \textit{SoftmaxConf} uses maximum softmax probability with uncertainty as $1 - \max(p_i)$. \textit{DeepEnsemble} utilizes disagreement between multiple networks. \textit{MCDropout} applies Monte Carlo dropout sampling for predictive variance estimation.

\begin{table}[h]
\begin{center}
\caption{EDL Ablation Study Results on PU - PC Task}
\label{tab:edl_ablation}
\small{
\begin{tabular}{lccc}
\toprule
\textbf{Variant} & \textbf{OS* (\%)} & \textbf{Unk (\%)} & \textbf{HOS (\%)} \\
\midrule
Original & 66.80 & \textcolor{red}{\textbf{91.90}} & \textcolor{red}{\textbf{77.38}} \\
TempScaling & 76.47 & \textcolor{blue}{\textbf{30.31}} & \textcolor{blue}{\textbf{43.41}} \\
Entropy & 76.11 & 21.26 & 33.24 \\
SoftmaxConf & \textcolor{blue}{\textbf{76.96}} & 20.91 & 32.88 \\
DeepEnsemble & \textcolor{red}{\textbf{79.51}} & 15.94 & 26.55 \\
MCDropout & 76.94 & 15.89 & 26.33 \\
\bottomrule
\end{tabular}}
\end{center}
\end{table}

The results show EDL's superiority for OSDG, achieving the highest HOS score and excellent unknown class detection levels. While some alternatives like \textit{DeepEnsemble} and \textit{SoftmaxConf} achieve higher known class accuracy, they fail dramatically at unknown detection with poor rejection rates. The best alternative, \textit{TempScaling}, still underperforms EDL by a great margin. EDL's superiority stems presumably from its principled Dirichlet-based uncertainty quantification, which naturally separates known from unknown classes under domain shift. Traditional methods like \textit{Entropy} and \textit{SoftmaxConf} lack the theoretical foundation to handle distributional uncertainty, while ensemble methods (\textit{DeepEnsemble}, \textit{MCDropout}) cannot capture the specific uncertainty patterns required for cross-domain hyperspectral classification. Overall, the original EDL framework shows optimal balance between known class performance and unknown detection, validating its effectiveness for open-set domain generalization in hyperspectral imaging applications.

\textbf{SSUD:}
Six ablation experiments were designed to evaluate each SSUD component (Table \ref{tab:ssud_ablation}). \textit{FixedWeights} replaces adaptive pathway weighting with fixed weights $w_1=w_2=0.5$ to assess the importance of learned fusion. \textit{NoDecoupling} always uses combined pathway uncertainty. \textit{SimpleAverage} employs arithmetic mean between spatial and spectral uncertainty. \textit{MaxUncertainty} chooses the maximum value between spatial and spectral uncertainty. \textit{NoConfidence} removes the confidence term from rejection scoring. \textit{NoUncertainty} eliminates uncertainty weighting, relying solely on confidence.

\begin{table}[h]
\begin{center}
\caption{SSUD Ablation Study Results on PU - PC Task}
\label{tab:ssud_ablation}
\small{
\begin{tabular}{lccc}
\toprule
\textbf{Variant} & \textbf{OS* (\%)} & \textbf{Unk (\%)} & \textbf{HOS (\%)} \\
\midrule
Original & 66.80 & \textcolor{red}{\textbf{91.90}} & \textcolor{red}{\textbf{77.38}} \\
NoDecoupling & 66.28 & \textcolor{blue}{\textbf{81.26}} & \textcolor{blue}{\textbf{72.96}} \\
FixedWeights & \textcolor{blue}{\textbf{73.99}} & 46.10 & 56.81 \\
MaxUncertainty & 69.01 & 27.45 & 39.28 \\
NoConfidence & 72.02 & 22.44 & 34.22 \\
SimpleAverage & 70.96 & 19.71 & 30.85 \\
NoUncertainty & \textcolor{red}{\textbf{77.25}} & 8.63 & 15.53 \\
\bottomrule
\end{tabular}}
\end{center}
\end{table}

\begin{figure*}[h]
  \begin{center}
  \subfloat[Impact of hyperparameter $\alpha$.]{\label{fig.param.alpha}\includegraphics[width=0.29\textwidth]{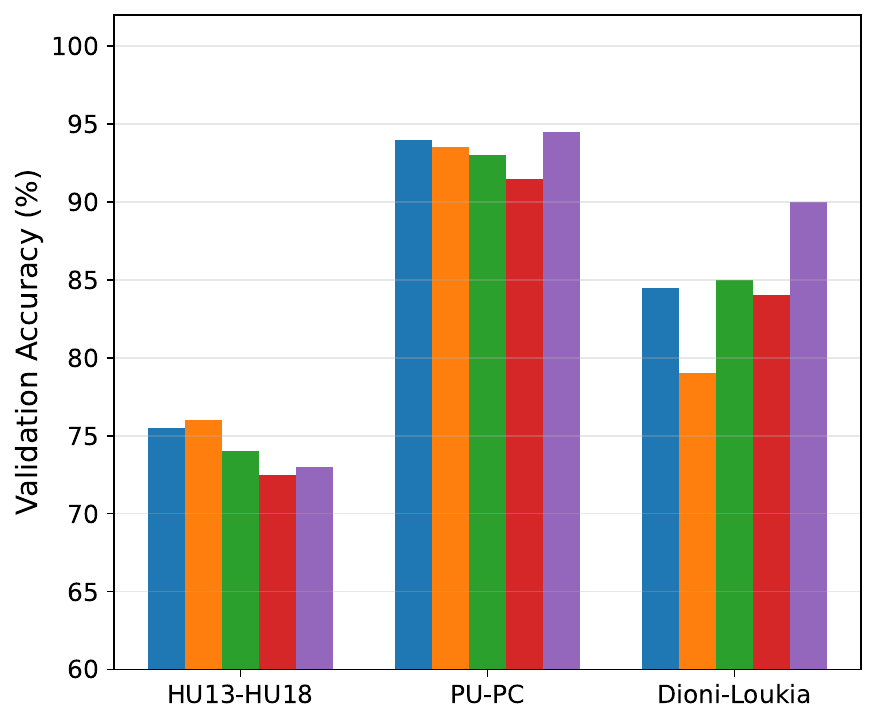}} \quad
   \subfloat[Impact of hyperparameter $\beta$. ]{\label{fig.param.beta}\includegraphics[width=0.29\textwidth]{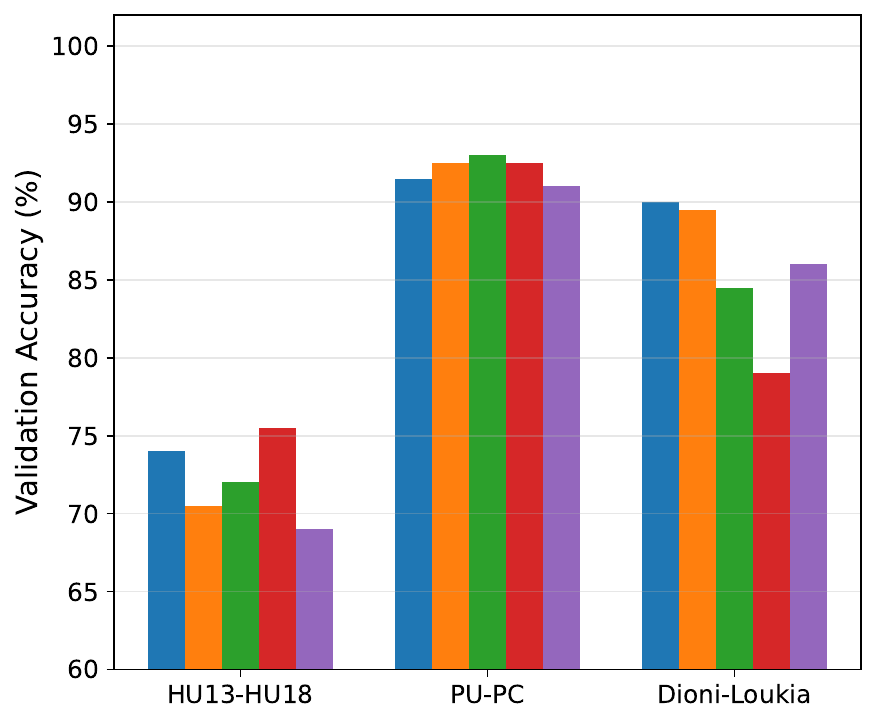}} \quad
  \subfloat[Impact of hyperparameter $\gamma$.]{\label{fig.param.gamma}\includegraphics[width=0.29\textwidth]{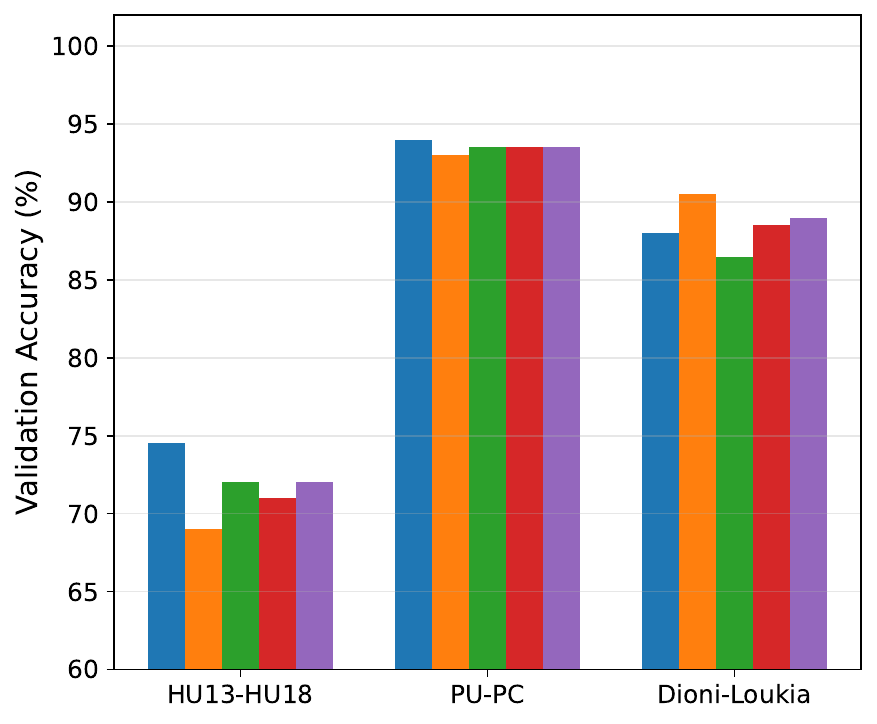}}\

\caption{The grid search method is used to find the optimal values of hyperparameters $\gamma$, $\beta$, and $\alpha$.
\crule[gamma01]{0.5cm}{0.25cm} Parameter value = 0.1, 
\crule[gamma03]{0.5cm}{0.25cm} Parameter value = 0.3,
\crule[gamma05]{0.5cm}{0.25cm} Parameter value = 0.5,
\crule[gamma07]{0.5cm}{0.25cm} Parameter value = 0.7,
\crule[gamma09]{0.5cm}{0.25cm} Parameter value = 0.9}
  \label{fig:parameter-tuning}
  \end{center}
\end{figure*}

\begin{table*}[h]
\begin{center}
\caption{Computation Time (s), FLOPs, and Parameters for Each Task With Different Methods.}
\label{tab:computation_results}
\small{
\begin{tabular}{l|l|ccccccccc|c}
\hline
\multirow{2}{*}{Dataset} & \multirow{2}{*}{Metric} & \multicolumn{9}{c|}{Domain Adaptation} & DG \\
\cline{3-12}
& & DACD & OSBP & DAMC & STA & MTS & UADAL & OMEGA & ANNA & WGDT & Ours \\
\hline
\multirow{2}{*}{PU} & training time & 277.82 & 384.42 & 350.31 & 296.29 & 2060.77 & 2187.91 & 500.25 & 770.09 & 312.11 & 2280 \\
& testing time & 7.66 & 12.03 & 6.16 & 50.15 & 24.98 & 28.77 & 6.02 & 54.90 & 13.22 & 38.8 \\
\multirow{2}{*}{PC} & FLOPS & 203M & 199M & 730M & 26047M & 52112M & 5089M & 6301M & 56862M & 8394M & 568M \\
& parameters & 330K & 316K & 1439K & 27929K & 57368K & 48853K & 23525K & 32976K & 1976K & 56760K \\
\hline
\multirow{2}{*}{HU13} & training time & 242.41 & 421.22 & 307.04 & 308.98 & 2115.01 & 10117.76 & 635.01 & 679.24 & 263.86 & 576 \\
& testing time & 33.68 & 70.63 & 34.13 & 234.79 & 153.70 & 186.81 & 32.42 & 299.33 & 77.13 & 234 \\
\multirow{2}{*}{HU18}& FLOPS & 105M & 102M & 583M & 26028M & 52076M & 5076M & 6283M & 56860M & 3932M & 567M \\
& parameters & 322K & 308K & 1431K & 27928K & 57365K & 48850K & 23523K & 32975K & 1850K & 56730K \\
\hline
\multirow{2}{*}{Dioni} & training time & 296.61 & 603.00 & 367.35 & 317.41 & 2186.23 & 1009.43 & 611.95 & 997.27 & 605.11 & 738 \\
& testing time & 0.96 & 0.95 & 0.96 & 6.63 & 4.05 & 3.78 & 1.29 & 7.18 & 1.87 & 5.1 \\
\multirow{2}{*}{Loukia} & FLOPS & 336M & 333M & 931M & 26072M & 52939M & 5109M & 6637M & 56872M & 1451M & 568M \\
& parameters & 340K & 327K & 1452K & 27932K & 58424K & 48866K & 23531K & 32980K & 2150K & 56830K \\
\hline
\end{tabular}}
\end{center}
\end{table*}

\textit{NoDecoupling} achieves competitive performance, confirming that reliability-based pathway selection provides meaningful improvements but is not the dominant factor. \textit{FixedWeights} shows high known class accuracy but poor unknown rejection, revealing that adaptive pathway weighting is essential for uncertainty-driven open-set decisions. Methods that oversimplify uncertainty estimation perform poorly: \textit{SimpleAverage} and \textit{MaxUncertainty} show that sophisticated uncertainty fusion significantly outperforms naive approaches. \textit{NoConfidence} shows moderate degradation, indicating that confidence-based rejection scoring provides complementary information to uncertainty estimates. Most critically, \textit{NoUncertainty} achieves the highest known class accuracy but catastrophic unknown rejection failure, confirming that uncertainty quantification is indispensable for open-set recognition and cannot be replaced by confidence alone. The original SSUD design achieves optimal synthesis, validating that each component contributes meaningfully to the framework's superior open-set domain generalization performance.

\subsection{Parameter Discussion}
\label{sec:prams}
The effect of various hyperparameters on performance has been explored through grid search. Specifically, ablation studies were conducted on each of $\alpha$, $\beta$, and $\gamma$ by adjusting each parameter within the range $\{0.1, 0.3, 0.5, 0.7, 0.9\}$ while fixing the other two parameters. The best hyperparameters were selected based on validation performance.

\textbf{EDL Loss Weight }($\alpha$):
The EDL loss weight controls the emphasis on uncertainty quantification in the learning process. As shown in Fig.~\ref{fig.param.alpha}, all datasets exhibit a similar pattern, where accuracy initially improves with increased $\alpha$ but degrades when $\alpha$ becomes too large. This behavior suggests that moderate uncertainty regularization enhances generalization by preventing overconfident predictions, while excessive weighting may over-penalize the model. The model demonstrates moderate sensitivity to $\alpha$.

\textbf{Domain Adversarial Loss Weight }($\beta$):
The domain adversarial loss weight balances the strength of domain alignment. Fig.~\ref{fig.param.beta} shows dataset-dependent sensitivity patterns. The Houston dataset shows high sensitivity with a sharp peak at a higher hyperparameter value, suggesting strong domain shift that requires aggressive alignment. In contrast, Pavia exhibits lower sensitivity, indicating moderate domain discrepancy. The Dioni dataset achieves optimal performance with minimal alignment, implying relatively similar domains where excessive alignment may harm class discriminability.

\textbf{Reconstruction Loss Weight }($\gamma$):

The reconstruction loss weight balances feature extraction quality through the DCRN module. As illustrated in Fig.~\ref{fig.param.gamma}, the model generally favors lower $\gamma$ values, suggesting that excessive focus on reconstruction may divert the model from the primary classification objective. The model shows relatively low sensitivity to $\gamma$, with stable performance across lower $\gamma$ values and only modest degradation at higher values.

The analysis reveals that $\beta$ exhibits the highest dataset-specific sensitivity, while $\alpha$ and $\gamma$ demonstrate more consistent behavior across datasets.

\subsection{Computation analysis}
A computational analysis has been conducted to evaluate the complexity of the proposed DG method, in terms of training time, testing time, number of floating point operations (FLOPs), and number of parameters (Table \ref{tab:computation_results}).
It exhibits competitive computational efficiency across all datasets. Notably, it achieves moderately reduced training times compared to some DA methods, while maintaining reasonable testing times and parameter counts. The FLOPs of the proposed method are comparable to or lower than many DA approaches, and the parameter count falls within a reasonable range. These results show that the presented DG framework provides effective cross-domain performance.

\section{Conclusion}
\label{sec:con}
The proposed OSDG framework is a novel architecture for cross-scene hyperspectral image classification. Leveraging spectrum-invariant frequency disentanglement, dual-channel residual networks, evidential deep learning, and spectral-spatial uncertainty disentanglement, the framework effectively handles unknown classes in target domains while generalizing across unseen domains without target-specific adaptation. 
Experiments on three cross-scene HSI classification benchmarks showed the proposed approach achieved superior or competitive performance compared to state-of-the-art domain adaptation methods. Results showed high performance with effective unknown class rejection, making the proposed framework promising for practical HSI classification applications. 

Future work will focus on developing class-specific uncertainty thresholds to address the observed performance limitations on spectrally variable land cover types, and extending the framework to multi-source DG scenarios for enhanced robustness across diverse geographical regions.

\bibliographystyle{unsrt} 
\bibliography{ref}

\end{document}